\documentclass{article}

\usepackage{PRIMEarxiv}

\usepackage[utf8]{inputenc} 
\usepackage[T1]{fontenc}    
\usepackage{hyperref}       
\usepackage{url}            
\usepackage{booktabs}       
\usepackage{amsfonts}       
\usepackage{nicefrac}       
\usepackage{microtype}      
\usepackage{lipsum}
\usepackage{fancyhdr}       
\usepackage{graphicx}       
\usepackage{natbib}

\usepackage{algorithm}
\usepackage{algpseudocode}
\usepackage{adjustbox}
\usepackage{float}

\usepackage{amsmath,amsfonts,bm}

\def\eqref#1{equation~\ref{#1}}

\def\1{\bm{1}}

\DeclareMathAlphabet{\mathsfit}{\encodingdefault}{\sfdefault}{m}{sl}
\SetMathAlphabet{\mathsfit}{bold}{\encodingdefault}{\sfdefault}{bx}{n}

\usepackage{hyperref}
\usepackage{url}
\usepackage{booktabs}
\usepackage{comment}

\graphicspath{{media/}}     

\title{Robust Graph Representation Learning \\ via Predictive Coding}

\author{Billy Byiringiro$^1$, Tommaso Salvatori$^{2}$, Thomas Lukasiewicz$^{1,2}$ \\
  $^1$ Department of Computer Science,
  University of Oxford, UK \\
  $^2$ Institute of Logic and Computation, TU Wien, Austria \\
  \texttt{billy.byiringiro@cs.ox.ac.uk, tommaso.salvatori@tuwien.ac.at,} \\ \texttt{thomas.lukasiewicz@cs.ox.ac.uk} \\
}

\begin{document}
\maketitle


\begin{abstract}

Predictive coding is a message-passing framework initially developed to model information processing in the brain, and now also topic of research in machine learning due to some interesting properties. One of such properties is the natural ability of generative models to learn robust representations thanks to their peculiar credit assignment rule, that allows neural activities to converge to a solution before updating the synaptic weights. Graph neural networks are also message-passing models, which have recently shown outstanding results in diverse types of tasks in machine learning, providing interdisciplinary state-of-the-art performance on structured data. However, they are vulnerable to imperceptible adversarial attacks, and unfit for out-of-distribution generalization. In this work, we address this by building models that have the same structure of popular graph neural network architectures, but rely on the message-passing rule of predictive coding. Through an extensive set of experiments, we show that the proposed models are (i)~comparable to standard ones in terms of performance in both inductive and transductive tasks, 
(ii)~better calibrated,  and (iii)~robust against multiple kinds of adversarial attacks.

\end{abstract}

\section{Introduction}

Extracting information from structured data has always been an active area of research in machine learning. This, mixed with the rise of deep neural networks as the main model of the field, has led to the development of \emph{graph neural networks}  (GNNs). These models have achieved results in diverse types of tasks in machine learning, providing interdisciplinary state-of-the-art performance in areas such as e-commerce and financial fraud detection \citep{zhang2022efraudcom, wang2019semi}, drug and advanced material discovery \citep{bongini2021molecular, zhao2021csgnn, xiong2019pushing}, recommender systems \citep{wu2021self}, and social networks \citep{liao2018attributed}. Their power lies in a message passing mechanism among vertices of a graph, performed iteratively at different levels of hierarchy of a deep network. Popular examples of these models are \emph{graph convolutional networks} (GCNs) \citep{welling2016semi} and \emph{graph attention networks} \citep{velivckovic2017graph}. Despite the aforementioned results and performance obtained in the last years, these models have been shown to lack robustness. They are in fact vulnerable against carefully-crafted adversarial attacks \citep{zugner2018adversarial,gunnemann2022graph,dai2018adversarial,zugner2019adversarial} and unfit for out-of-distribution generalisation \citep{hu2020open}. This prevents GNNs from being used in critical tasks, where misleading predictions may lead to serious consequences, or maliciously manipulated signals may lead to the loss of a large amount of money. 

More generally, robustness has always been a problem of deep learning models, highlighted by the famous example of a panda picture being classified as a gibbon with almost perfect confidence after the addition of a small amount of adversarial noise \citep{akhtar2018threat}. To address this problem, an influential work has shown that it is possible to treat a classifier as an energy-based generative model, and train the joint distribution of a data point and its label to improve robustness and calibration \citep{grathwohl2019your}. Justified by this result, this work studies the robustness of GNNs trained using an energy-based training algorithm called \emph{predictive coding} (PC), originally developed to model information processing in hierarchical generative networks present in the neocortex \citep{rao1999predictive}. Despite not being initially developed to perform machine learning tasks, recent works have been analyzing possible applications of PC in deep learning. This is motivated by interesting properties of PC, as well as similarities with backpropagation~(BP) in terms of update of the parameters: when used to train classifiers, PC is able to approximate the weight update of BP on any neural network \citep{whittington2017approximation,millidge2021predictive}, and a variation of it is able to exactly replicate the weight update of BP \citep{song2020can,salvatori2022reverse}. It has been shown that PC is able to train powerful image classifiers \citep{he2016deep}, is able to perform generation tasks \citep{ororbia2022neural}, continual learning \citep{ororbia2020continual}, associative memories \citep{salvatori2021associative,tang2022recurrent}, reinforcement learning \citep{ororbia2022active}, and train neural networks with any structure \citep{salvatori2022learning}. It is, however, the unique credit assignment rule of predictive coding, where errors are dynamically redistributed throughout the network and concentrated where they are most needed before performing a weight update, which is interesting to us. It has in fact been shown that this allows PC models to perform better than standard ones in many biologically relevant scenarios \citep{song2022inferring}. In this work, we extend the study of PC to structured data, and show that PC is naturally able to train robust classifiers due to its energy-based formulation. To show that, we first show that PC is able to match the performance of BP on small and medium tasks, hence showing that the results on image classification \citep{whittington2017approximation} extend to graph data, and then showing the improved calibration and robustness against adversarial attacks of models trained this way. Summarizing, our contributions are briefly as follows:
\begin{itemize}
    \item We introduce and formalize a new class of message passing models, which we call \emph{graph predictive coding networks} (GPCN). We show that these models achieve a performance comparable with equivalent GCNs trained using BP in multiple tasks, and propose a general recipe to train any message-passing GNN with PC. 
    \item We empirically show that GPCNs are less confident in their prediction, and hence produce  models that are better calibrated than equivalent GCNs. Our results show large improvements in expected calibration error and maximumum calibration error on the CORA, CiteSeer, and PubMed datasets. This proves the ability of GPCNs to estimate the likelihood close to the true probability of a given data point and capacity to better capture uncertainty in its prediction.
    
    \item We further conduct an extensive robustness evaluation using advanced graph adversarial attacks on various dimensions: poisoning and evasion, global and targeted, and  direct and indirect. In these evaluations, (i) we introduce PC-based graph attention networks (PC-GATs), and we show that (ii) GPCNs outperform standard GCNs on all kinds of evasion attacks, (iii) GPCNs and PC-GATs outperform their counterpart on poisoning attacks and random-poisoning attacks on large datasets, and (iv) they naturally obtain a better performance on various datasets than other complex methods that use tricks designed to make the model more robust \citep{zhu2019robust}. Note that the goal of these experiments is not to provide state-of-the-art results, but to show that GPCNs have a natural predisposition towards learning robust representations.
\end{itemize}

\section{Preliminaries}

In this section, we review the general framework of message-passing neural networks (MPNNs) \citep{gilmer2017neural}. Assume a graph $G = (V,E, X)$ with a set of nodes $V$, a set of edges $E$, and a set of attributes or properties of each node in the graph, described by a matrix $X \in \mathbb{R}^{ |V| \times d}$. 
The idea behind MPNNs is to begin with certain initial node characteristics and iteratively modify them over the course of $k$ iterations using information gained from neighbours of each node.  The representation  $\mathbf{h}_{u}^{(k)}$ of a node $u \in V$ at layer $k$ is iteratively modified  as follows:
\begin{equation}
\label{eqn:ggn_eqn}
\mathbf{h}_{u}^{(k)}=\operatorname{update}^{(k)}\left(\mathbf{h}_{u}^{(k-1)}, \  \operatorname{aggregate}^{(t)}\left(\left\{\mathbf{h}_{v}^{(k-1)} \mid v \in N(u)\right\}\right)\right)\,,
\end{equation}
where $N(u)$ is a set of neighbors of node $u$, and \emph{update} and \emph{aggregate} are differentiable functions. The \emph{aggregate} function has to be a permutation-invariant to maintain symmetries  necessary when operating on graph data, such as locality and invariance properties. In this work, we mainly focus on graph convolutional networks  (GCNs). Here, the aggregation function is a weighted combination of neighbour characteristics with predetermined fixed weights, and the \emph{update} function is a linear transformation.

\subsection{Predictive Coding Graphs}
\label{sec:pc}

Predictive coding networks (PCNs) were first introduced for unsupervised feature learning \citep{rao1999predictive}, and later extended to supervised learning \citep{whittington2017approximation}. Here, we describe a recent formulation, called PC graphs, that allows to use PC to train on graphs with any topology \citep{salvatori2022learning}. What results, is a message passing mechanism that is similar to that of GNNs, but with no multilayer structure. Let us consider a directed graph $G=(V,E)$, where $V$ is a set of $n$ vertices, and $E$ the set of directed edges. Every vertex $u$ is equipped with a value node $\mathbf h_{u,t}$, which denotes the neural activity of the node $u$ at time $t$, and is a variable of the model, and every edge has a weight $w_{u,v}$. Every node has a \emph{prediction} $\mu_{u,t}$, given by the incoming signals from other layers processed by an aggregation function (in practice, always the sum), and prediction error $\varepsilon_{u,t}$, given by the difference between the real value of a node and its prediction. In detail, 
\begin{equation}
\label{eqn:error_update}
\mu_{u,t}=\sum_{v \in p(u)} w_{v,u} f \left(\mathbf h_{v,t} \right) \ \ \text{ and } \ \ \varepsilon_{u,t}= \mathbf h_{u,t}-\mu_{u,t} \text {, }
\end{equation}
where $p(u)$ denotes the set parent nodes of $u$. As PC graphs are energy-based models, training happens through minimization of the global energy in each layer. This global energy $F_{t}$ is the sum of the prediction errors of the network:
\begin{align}
F_{t}=\frac{1}{2} \sum_{u}\left(\varepsilon_{u,t}\right)^{2}.
\label{eq:energy}
\end{align}
%
Learning happens in two phases, called inference and weight update. The inference phase is a message passing process, where the weights are fixed, and the values are continuously updated according to neighbour information. Differently from GNNs, where the update rule is given, here it follows an objective, that is to minimize the energy of Equation \ref{eq:energy}. This is done via gradient descent until convergence. The update rule is the following:
\begin{align}
\Delta{ \mathbf h}_{u,t} \sim \partial \mathcal{E}_t/\partial x_{u,t} =  -\varepsilon_{u,t} + f' (\mathbf h_{u,t} ) \sum_{v \in c(u)} \varepsilon_{v,t} w_{v,u},
\label{eq:x_update}
\end{align}
where $c(u)$ is the set of children vertices of $u$. To perform a weight update, we fix all the value nodes, and update the weights for one iteration by minimizing the same energy function via gradient descent as follows:
\begin{align}\label{eq:w_update}
\Delta w_{i,j}  \sim {\partial \mathcal{E}_t}/{\partial \theta_{i,j}} =  \alpha\cdot \varepsilon^l_{i,t} f (\mathbf h_{j,t})\,.
\end{align}

\section{Graph Predictive Coding Networks}

We now propose \emph{graph predictive coding networks} (GPCNs), obtained by using the multilayer structure of GNNs, and the learning mechanism of PC. To design GPCNs, we introduce two different message passing rules, and incorporate them inside the same training algorithm. Both rules are derived by minimizing the same energy function of Equation \ref{eq:energy} via gradient descent. Here, the PC graph has an hierarchical structure and intra-layer and inter-layer operations.  Inter-layer operations update neural activities and prediction according to information coming from the layer above, while intra-layer ones do it accordingly to the predictions computed from neighbour nodes according to an  aggregation mechanism.

\textbf{Inter-layer operation:} We follow the formulation of PC graphs that we described in Section~\ref{sec:pc}. For a node $\mathbf{u} \in V$ and a message passing layer $k$, we have neural activity state denoted as $\mathbf{h}^{k}_{u, t}$ and corresponding prediction-error state,  $\mathbf{\varepsilon}^{k}_{u, t}$. $t$ denotes the inference phase time step during energy minimization.  The predicted representation,  $\mathbf{\mu_{u, t}^{k}}$, at layer $k$ is calculated as follows:
\begin{equation}
\label{eqn:prediction_error_inter_layer}
\mathbf{\mu_{i, t}^{k}}=\operatorname{update}^{(k)}\left(\mathbf{h}_{u}^{(k-1)}, \text { aggregate }^{(t)}\left(\left\{\mathbf{h}_{v}^{(k-1)} \mid v \in N(u)\right\}\right)\right) \text { and } \mathbf{\varepsilon}_{u, t}^{k}=\mathbf{h}_{u, t}^{k} - \mathbf{\mu}_{u, t}^{k}\,.
\end{equation}
%
%
The embeddings of the nodes $u$ are obtained through global energy minimization during inference stage $\mathbf{h}^{k}_{i, t}$ as described in Section~\ref{sec:pc}.  

\textbf{Intra-layer operation:}
For intra-layer operations, we similarly apply the predictive coding mechanism to neighbourhood aggregation stage, where the neural activity state of neighboring nodes of node $u$ is denoted as $\mathbf{h}_{N(u)}^{(k)}$ and its corresponding prediction-error state and predicted state are denoted as $\mathbf{\varepsilon_{agg}}^{k}_{u, t}$ and $\mathbf{\mu_{agg}}^{k}_{u, t}$, respectively. The equations governing the dynamic of this model are the following:
\begin{align}
h_{u,t}^{k} & =\operatorname{update}^{(k)}\left(\mathbf{h}_{u}^{(k-1)}, \mathbf{h}_{N(u)}^{(k)}\right) \\ 
\mathbf{\mu_{agg}}^{k}_{u, t} & = \text { aggregate }^{(k)}\left(\left\{\mathbf{h}_{v}^{(k-1)} \mid v \in N(u)\right\}\right) \\
\mathbf{\varepsilon_{agg}}^{k}_{u, t} & =\mathbf{h}_{N(u)}^{(k)}-\mathbf{\mu_{agg}}^{k}_{u, t}\,.
\end{align}
In a similar fashion, $\mathbf{h}_{N(u)}^{(k)}$ is not updated directly, rather, it is updated during inference stage. In what follows, we test GPCNs on some standard benchmarks on both inductive and transductive tasks.

\begin{table}[t]
    \caption{Test accuracy on transductive tasks.}
    
    \smallskip 
    \centering
\begin{tabular}{@{}lcccc@{}}
\toprule
Method & CORA & CiterSeer & PubMed \\
\toprule

GCN & $\mathbf{80.72\pm1.05\%}$   & $67.12\pm1.53\%$ & $\mathbf{77.1\pm1.45\%}$\\
GPCN & $80.7\pm1.09\%$ & $\mathbf{67.26\pm1.28\%}$ & $76.2\pm2.44\%$\\
\bottomrule
\label{tab:trans}
\end{tabular}

\caption{F-1 score on inductive tasks.}

\smallskip 
\begin{tabular}{@{}lcccccc@{}}
\toprule
Method & CORA & CiterSeer & PubMed & PPI(Sup.) & PPI(Unsup.) \\
\toprule
GCN & $\mathbf{80\pm0.41\%}$   & $67.64\pm1.14\%$ & $77.0\pm0.46\%$ & $76.45\pm0.39\%$ & $52.44\pm0.37\%$\\
GPCN & $79.66\pm0.75\%$ & $\mathbf{69.68\pm0.37\%}$ & $\mathbf{77.12\pm0.47\%}$ & $\mathbf{78.31\pm0.47\%}$ & $\mathbf{54.41\pm 0.31\%}$\\
\bottomrule
\label{tab:ind}
\end{tabular}

\end{table}


\section{Experiments}

In this section, we perform extensive experiments to assess the performance of our proposed method on common benchmark tasks, their calibration, and the ability of our energy-based models to counter graph adversarial attacks. More details on the experimental setup, a description to reproduce all the results presented in this section, as well as further results, are given in the supplementary material. We now provide some information about the models and datasets used, as well as a description of the baselines that we will compare against. 

\textbf{Datasets.} Following many related works \citep{welling2016semi, velivckovic2017graph, zugner2018adversarial, zhu2019robust}, we conduct experiments using the standard citation graph benchmark datasets: CORA, CiteSeer,  and PubMed. We also employ the inherently inductive large-scale protein-to-protein interaction (PPI)~\citep{hamilton2017inductive, velivckovic2017graph} dataset to validate the scalability of our model. The dataset statistics are summarised in Table~\ref{tab:datasets} in the supplementary material.


\textbf{Baselines.} To evaluate the performance of our framework, we first compare GPCNs against GCNs~\citep{welling2016semi}, as our proposed method is based on the same message-passing scheme. To make sure that the comparison is as fair as possible, all the  models are identical in structure (number of parameters, depth, and width), and are  trained without dropout or batch-normalisation. Furthermore, in some tasks, we have also trained PC-based graph attention networks (PC-GATs), and compared against the standard formulation trained with backpropagation \citep{velivckovic2017graph}. We also refer to results from  Robust-GCN (RGCN)~\citep{zhu2019robust}, a popular model specifically developed to increase the robustness of GCNs.

\textbf{Evaluation metrics.}
We are interested in learning calibrated models, that is, the ability of a model to produce probability estimations that are accurate reflections of the correct likelihood of an event. To do that, we employ scalar quantification metrics, such as expected calibration error (ECE) and maximum calibration error (MCE). The first captures the notion of average miscalibration in a single number, and can be obtained by the expected difference between the accuracy and the confidence of a model, while the second quantifies worst-case expected miscalibration.

\subsection{Experiments on General Performance}
To study where we stand in terms of performance against standard GNNs, we now test our newly proposed model against GCNs trained with BP on both transductive and inductive tasks. As shown in Tables~\ref{tab:trans} and \ref{tab:ind}, our method is always comparable to the baseline in terms of performance. This is important, as it allows to make the improvements in robustness meaningful, which is the main goal of our work.

\subsection{Calibration Analysis}
\label{sec:cal_analysis}

\begin{figure}[t]
    \centering
    \includegraphics[width=\textwidth]{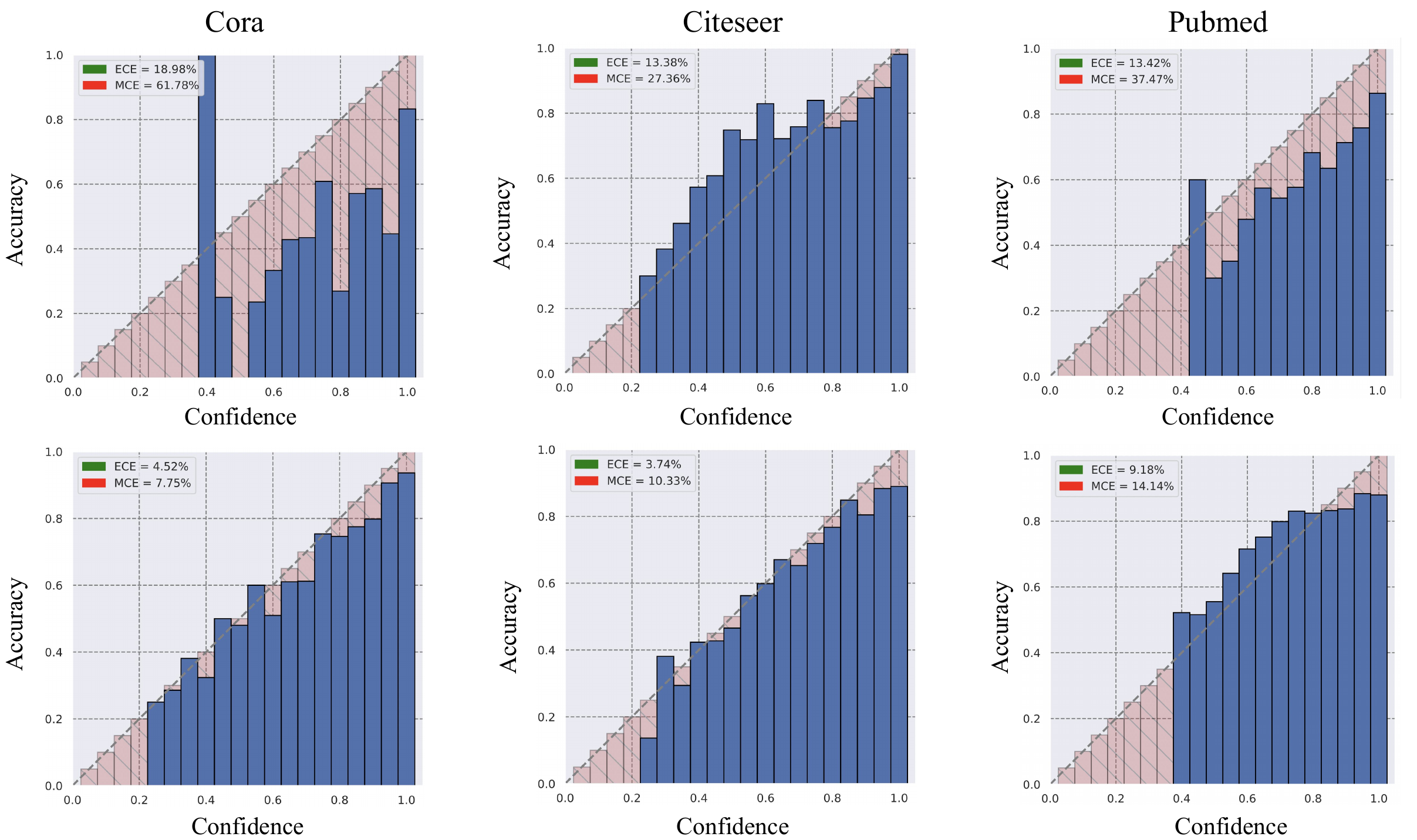}%
    \caption{Reliability diagrams on multiple datasets. In the top row, we have GCNs, and in the bottom row, GPCNs. Inside every figure, we have the MCE and ECE of every model (the lower the better). Any deviation from the diagonal line indicates miscalibration. In all cases, PC outperforms standard GCNs trained with BP. }%
    \label{fig:reliability}%
\end{figure}

Here, we investigate the calibration robustness of our GPCNs in comparison to GCNs. To do that, we use reliability diagrams introduced in \citep{guo2017calibration}. Reliability diagrams are a visual representation of model calibration that plots the expected sample accuracy as a function of confidence.  That is, if a model is perfectly calibrated, the diagrams would plot an identity function, while any variation implies miscalibration. The reliability diagrams of both GCNs and GPCNs are in  Fig.~\ref{fig:reliability}. GCNs are highly overconfident on most prediction confidence levels on the CORA dataset, and highly under-confident on 0.4 prediction confidence. Conversely, on CiteSeer, GCNs tend to be under-confident on most predictions, which proves the miscalibration of GCN models. Our model, on the other hand, tends to approximate a perfect calibration with very small variations on both CORA and CiteSeer. A  similar trend is seen on PubMed, and further evidence is supported by histograms of prediction confidence distributions provided in the supplementary material.

We have also quantified the calibration error using ECE and MCE. The results, reported inside the plots in Fig.~\ref{fig:reliability}, again show that our models are better calibrated than GCNs. These results are interesting, as they show that our models can effectively quantify uncertainty, which is crucial 
in highly critical settings. As different learning rates have different impact on calibration \citep{guo2017calibration}, in the supplementary material, we have  provided plots of how ECE and MCE change over time with different learning rates, as well as further details on the experiments. 


\begin{table}[t]
\caption{Robustness against perturbations.}
\centering
\begin{tabular}{@{}lcccc@{}}
\toprule
 & GCN~\citep{chen2021understanding} & RGCN~\citep{chen2021understanding} & GCPN \\
\toprule
CORA & $2.05\pm0.07$   & $2.79\pm0.10$ & $\mathbf{3.26\pm0.18}$\\
CiterSeer & $1.98\pm0.12$   & $2.02\pm0.23$ & $\mathbf{2.73\pm0.08}$\\
PubMed & $1.14\pm0.02$   & $1.48\pm0.02$ & $\mathbf{4.21\pm0.32}$\\
\bottomrule
\end{tabular}
\label{tab:direct_attack_tab}

\end{table}

\subsection{Evasion Attacks with Nettack}
\label{sec:evasion}

\begin{figure}[t]
    \centering
    \includegraphics[width=\textwidth]{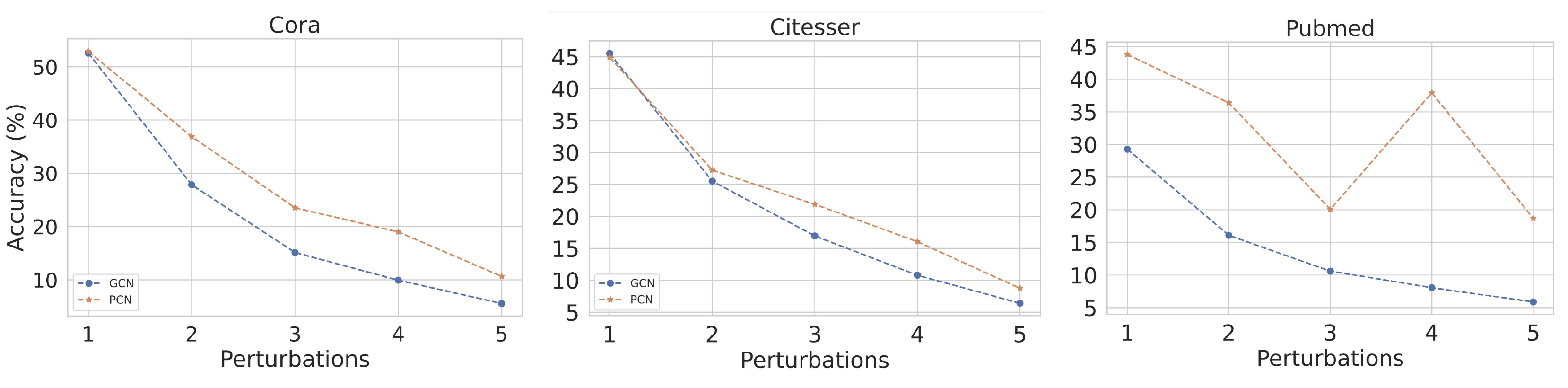}%
    \caption{Robustness against direct attacks with respect to the number of
perturbations on CORA (left), CiteSeer (centre), and PubMed (right). In orange, GPCNs, and in blue, GCNs.}%
    \label{fig:nettack_pertubation_plots}%
\end{figure}

We now evaluate our model on one of the advanced graph adversarial evasion attacks, Nettack \citep{zugner2019adversarial}. In targeted evasion attacks, the model parameters are kept fixed. Then, we employ Nettack, which uses a surrogate model trained on the same training set to attack selected victim nodes. Here, all experiments are repeated five times under different random~seeds.

Following the experimental setting of \citep{chen2021understanding}, we assess the robustness against structural attacks. Here, we randomly select $1000$ victims nodes from both the validation and the test set. As in previous works \citep{zugner2019adversarial, jin2020graph}, the perturbations budget ranges from $1$ to $5$, and each victim nodes is attacked separately. We employ $\sum_{q=1}^5 q \times p_q$ as holistic robustness metric, where $q$ denotes the number of perturbations, and $p_q$ is the classification accuracy corresponding to the perturbation budget $q$ \citep{chen2021understanding}. A larger value of this metric corresponds to higher robustness. The results are displayed in Table~\ref{tab:direct_attack_tab}, where GPCNs outperform GCNs and R-GCNs with drop-out and batch-normalisation. Interestingly, the highest robustness is achieved on PubMed, the largest dataset among the three. The result on each perturbation budget are also plotted in Fig.~\ref{fig:nettack_pertubation_plots}, and from the figure, we observe that GPCNs outperform GCNs for all budgets, especially for large numbers of perturbations.

\begin{figure}[t]
    \centering
    \makebox[\textwidth]{
    \includegraphics[width=\textwidth]{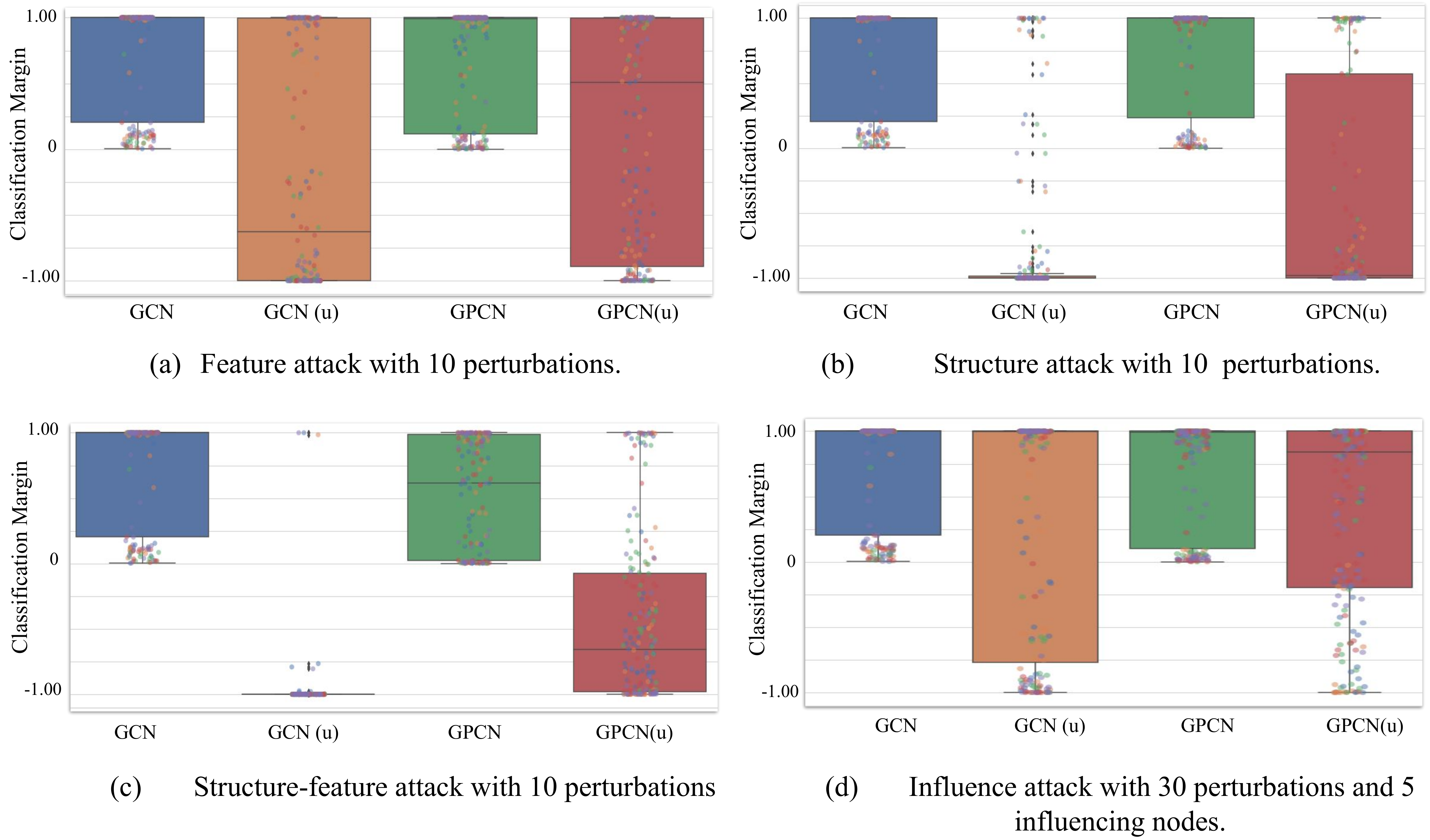}
    }
    \caption{Classification margin diagrams on different types of attacks on CORA. GCN(u) and GPCN(u) labels indicate the results of GCN and GPCN on clean/untempered graph data. Figure (a) corresponds to adversarial attacks on node feature attacks; Figure (b) corresponds to adversarial attacks on graph structures; Figure (c) corresponds to adversarial attacks on both structure and features; and  Figure (d) corresponds to indirect attacks where adversarial attack targets neighboring nodes of a victim node}
    \label{fig:e_attacks}
\end{figure}

We have also performed a semi-qualitative analysis of robustness using classification margin and box-plots. Here, victim nodes are selected in similar fashion as in the Nettack paper \citep{zugner2019adversarial}: we have  selected $10$ nodes with highest classification margins, $10$ other nodes with lowest classification margins, and $20$ nodes that are randomly selected from the test set. We then perform a range of attacks such as direct and indirect attacks and feature or/and structure attacks, and evaluate the robustness on varying perturbation budgets. In the box-plots represented in Fig.~\ref{fig:e_attacks}, each point represents one victim node, and the color of each point indicates the random seed on which an experiment is performed. The suffix '(u)' indicates the performance of a model on clean graphs. A more robust model is one that retains  higher classification margins after an attack.

\begin{enumerate}
    \item \textbf{Structure and feature attack:} Figure~\ref{fig:e_attacks} (c) shows that with only 10 perturbations on the neighbourhood structure and features of victim nodes, the classification margin of victim nodes collapses to $-{1}$ on GCNs, while GPCNs stay relatively robust with many victim nodes retaining positive classification margins, meaning that they were not adversarial affected by the attack. This trend is reflected for all numbers of perturbations ($2,5,10$), as reported in the supplementary material. In particular, when the number of perturbations is 2, the median of classification margin for GCNs fall closer to $-{1}$, while GPCNs protectaround 50 percent of the victim nodes retaining a positive classification margin after the attack.

    \item \textbf{Feature attacks:} Since feature attacks do not affect GNNs as much as structure attacks, we use a high perturbation budget for features with perturbation numbers in $\{1, 5, 10, 30, 50, 100\}$. We observe a similar trend as above, where a small perturbation on features does not affect the model much. However, when the perturbation rate becomes large, GPCNs are much better in resisting the attacks. In detail, when the perturbation rate is equal to 30 (see Fig.~\ref{fig:e_attacks} (a)),  GCNs misclassify around $70\%$ of the victim nodes, while GPCNs less than $30\%$. In the supplementary material we show that when the perturbation budget is increased to $100$, GCNs mislassify all victim nodes, while GPCNs are still able to correctly classify most victim nodes in the upper quartile.
    
    \item \textbf{Structure attacks}: GPCNs also consistently outperform GCNs under structure-only attacks, as it can be observed in Fig.~\ref{fig:e_attacks} (b) under 10 perturbations. More figures under different numbers of perturbations ($1,2,3,5$), which show similar results,   are provided in supplementary material.

    \item \textbf{Indirect Attacks:} For indirect attacks, we choose 5 influencing/neighboring nodes to attack for each victims node. Again, GPCNs consistently outperform GCNs. Interestingly, we observed that GPCNs are hardly affected on all perturbation budgets as the lower quartiles of all box plots stay in the positive half of classification margin for all attacks, as shown in the supplementary material.

\end{enumerate}


\subsection{Global Poisoning Attacks using Meta Learning (Mettack)}
\label{sec:mettack}

Finally, we perform global poisoning attacks using the Mettack technique \citep{zugner2019adversarial}. In poisoning attacks, only the training data are corrupted and are tempered with in a manner that renders the target model fail to learn. This is the most common type of graph attacks in the real world, as malicious individuals can change the training data, but do not have access to the parameters of the model  \citep{jin2021adversarial}. As Mettack has several variants, we use the same setting as in the Pro-GNN paper \citep{jin2020graph} and employ the most destructive variant known as Meta-Self on CORA and CiteSeer, and apply A-Meta-Self (approximate faster version of Meta-Self) on PubMed due to computation limitations. The perturbation rate is varied from $0\%$ to $25\%$ with a step of 5, and the results are reported in Table~\ref{table:mettack}, where we also compare with the results obtained in the Pro-GNN work. Note that the reason for the accuracy for $0$ perturbations to be different from the one we reported earlier, is that here we only use the largest component of a graph instead of using all nodes. As it can be seen from Table~\ref{table:mettack}, GPCNs consistently perform better than GCNs on all datasets with more than $10\%$ increase in robustness on CORA when the perturbation rate is  $25\%$. GPCNs and PC-GATs also outperform other methods under various perturbation rates on PubMed, the most challenging dataset, with more than $7\%$ improvement over both GATs and RGCNs when the perturbation rate is $25\%$. In the supplementary material, we report similar results for global random attacks and evasion attacks.

\begin{table}[t]
    \caption{Classification performance of the models under poisoning global attack with Metattack. }
  \label{table:mettack}    
\begin{adjustbox} {max width=\textwidth}

\begin{tabular}{ cc|ccccc}

\multicolumn{7}{c}{\textbf{\textit{Poisoning Attack with Mettack}}} \\
\hline
\textbf{Dataset} & Ptb Rate (\%)&  \textbf{GCN} & \textbf{GPCN-GCN} & \textbf{GAT \citep{jin2020graph}}& \textbf{GPCN-GAT} & \textbf{RGCN \citep{jin2020graph}}  \\
\hline
     & $0$ & $82.87\pm0.75$   & $83.17\pm0.78$ & $\mathbf{83.97\pm0.65}$ & $82.8{\pm1.21}$ & $83.09\pm0.44$\\
     & $5$ & $76.4\pm0.88$   & $78.17\pm1.13$ & $\mathbf{80.44\pm0.74}$ & ${78.91\pm0.71}$ & $77.42\pm0.39$\\
CORA & $10$ & $67.98\pm0.99$   & $71.21\pm1.13$ & $\mathbf{75.61\pm0.59}$ & ${72.22\pm1.21}$ & $72.22\pm0.38$  \\
     & $15$ & $60.3\pm1.73$   & $65.14\pm1.84$ & $\mathbf{69.78\pm1.28}$ & ${66.77\pm1.15}$ & $66.82\pm0.39$\\
     & $20$ & $50.31\pm1.69$   & $55.83\pm3.39$ & ${59.94\pm0.92}$ & ${54.7\pm1.34}$ & $\mathbf{59.27\pm0.37}$\\
     & $25$ & $44.16\pm0.88$   & $54.27\pm7.25$ &  $\mathbf{54.78\pm0.74}$& ${49.68\pm1.08}$ & $50.51\pm0.78$\\
\hline
     & $0$ & $72.43\pm0.05$   & $72.64\pm0.51$ & $73.26\pm0.83$ & $\mathbf{73.58\pm0.13}$ & $71.20\pm0.83$\\
     & $5$ & $71.53\pm0.43$   & $72.1\pm1.07$ & $\mathbf{72.89\pm0.83}$ & $72.61\pm0.57$ & $70.50\pm0.43$\\
CiteSeer & $10$ & $68.59\pm0.65$   & $ 69.04\pm0.45 $ & $\mathbf{70.63\pm0.48}$ & $70.34\pm0.25$ & $67.71\pm0.30$ \\
     & $15$ & $65.02\pm1.16$   & $ 65.95\pm1.21 $ & $\mathbf{69.02\pm1.09}$ &  $68.12\pm0.28$ & $65.69\pm0.37$\\
     & $20$ & $53.77\pm0.73$   & $ 56.73\pm1.67 $ & $61.04\pm1.52$ & ${59.12\pm0.84}$ & $\mathbf{62.49\pm1.22}$\\
     & $25$ & $57.49\pm2.13$   & $ 58.43\pm1.7 $ & ${61.85\pm1.12}$ & $\mathbf{62.02\pm0.73}$ & $55.35\pm0.66$\\
\hline
     & $0$ & $85.37\pm0.06$   & $85.3\pm0.3$ & $83.73\pm0.40$ & ${84.56\pm0.22}$ & $\mathbf{86.16\pm0.18}$ \\
     & $5$ & $81.4\pm0.12$   & $\mathbf{81.99\pm0.24}$ & $78.00\pm0.44$ & ${81.16\pm0.36}$ & $81.08\pm0.20$ \\
PubMed & $10$ & $79.73\pm0.27$   & $\mathbf{80.47\pm0.73}$ & $74.93\pm0.38$ & ${76.7\pm0.74}$ & $77.51\pm0.27$ \\
     & $15$ & $77.03\pm0.11$   & $\mathbf{79.38\pm0.43}$ & $71.13\pm0.51$ & ${73.87\pm0.62}$ & $73.91\pm0.25$ \\
     & $20$ & $75.59\pm0.26$   & $\mathbf{78.01\pm0.37} $ & $68.21\pm0.96$ & ${72.14\pm0.54}$ & $71.18\pm0.31$ \\
     & $25$ & $73.34\pm0.19$   & $\mathbf{75.69\pm1.43}$ & $65.41\pm0.77$ & ${68.94\pm0.6}$ & $67.95\pm0.15$\\
\hline
\end{tabular}

 \end{adjustbox}
 
\end{table}

\section{Related Work}

\textbf{Adversarial attacks on graphs.} The recent revelation of lacks of robustness of the current graph learning methods inspired a body of work that attempts to enhance the robustness of graph machine learning. Those techniques can generally be classified into three categories: robust representation, robust detection, and robust optimisation \citep{ma2021understanding}. Robust representation entails techniques that seek to map a graph representation into a resilient embedding space by minimising the loss objective function of anticipated worst-case perturbation approximations such as robustness certificates \citep{bojchevski2019certifiable} and known adversarial samples \citep{xu2019topology}. Robust detection techniques, on the other hand, recognise that the dearth of robustness of GNNs stems from the local message-passing aggregation phase; thus, they selectively choose which neighbourhood nodes to include in the aggregation based on some properties. Popular techniques in this category include the Jaccard method \citep{wu2019adversarial}, which removes edges of some nodes whose Jaccard similarity is below a certain threshold, and the singular value decomposition method \citep{entezari2020all}, which preprocesses a graph by generating a low-rank approximation of it. Finally, robust optimisation is concerned with regularisation techniques that avoid extreme embeddings.  GCN-LFR (Low-Frequency based Regularisation) \citep{chang2021not} adopts a robust co-training paradigm that derive the robustness from the eligible low-frequency components, while MedianGCN \citep{chen2021understanding} leverages robust aggregation functions (i.e., the median and trimmed mean) that ignore outliers based on a breakdown point characterisation. MedianGCN is very similar to SMGCN \citep{geisler2020reliable}, which introduces the soft medoid function as a message-aggregation method to produce a robust representation. Robust-GCN (RGCN)  \citep{zhu2019robust} embeds a node representation as a Gaussian distribution and utilises a variance-based attention mechanism during the neighbourhood message aggregation phase.


\textbf{Energy-based Models (EBMs).} Although there is a considerable interest in integrating the energy-based view into deep learning \citep{xie2016theory,nijkamp2019learning, grathwohl2019your, song2021train}, only a handful of works have transferred it to graph machine learning \citep{di2022graph}. Here, most lines of research have largely concentrated on graph generation tasks with models such as GNN-EBMS \citep{liu2020graph} and GraphEBM \citep{liu2021graphebm}.  Only one nascent work has recently attempted to expand the GCN classifier to an energy-based model named GCN-JEMO \citep{shin2021energy}. GCN-JEMO derives its energy from graph properties and was demonstrated to achieve a comparable discriminative performance to classic GCN but with increased robustness. In contrast to our work, GCN-JEMO relies on a non-standard training method and is not tested on adversarial attacks. 

\textbf{Predictive coding.} Recently, many works have been developed that use PC to address machine learning problems. A first example is computer vision, where recent works have performed image classification with simple experiments on MNIST \citep{whittington2017approximation}, or more complex ones on ImageNet \citep{he2016deep}. Other examples are image generation \citep{ororbia2022neural}, associative memories \citep{salvatori2021associative}, continual learning \citep{ororbia2020continual}, reinforcement learning \citep{ororbia2022active}, and NLP \citep{pinchetti2022}. We conclude by referring to a  more theoretical direction, that is,  Friston's free energy principle and active inference \citep{friston2010,friston2006free,friston2016active}.

\section{Summary and Outlook}

In this work, we have explored a new framework to perform machine learning on structured data, inspired from the neuroscience theory of predictive coding. First, we have defined the model, and then we have shown that it is able to reach competitive performance in both inductive and transductive tasks, with respect to similar models trained with BP. We have then tested this framework on robustness tasks, with extensive results showing that simply training GNNs using PC instead of BP, results in models that are better calibrated, and more robust against adversarial attacks. As we have used the original formulation adapted to GNNs, with no further effort put in increasing the robustness of the trained models, future work should focus on scaling up the results of this paper to large-scale models, and research on variations of the proposed framework that make these models even more robust. More generally, this work shrinks the gap between computational neuroscience and machine learning, by showing that biologically plausible methods are able to reach competitive performance on complex~tasks.

\section*{Acknowledgments}
This work was supported by the Alan Turing Institute under the EPSRC grant EP/N510129/1, 
by the AXA Research Fund, the EPSRC grant EP/R013667/1, the MRC grant MC\textunderscore UU\textunderscore 00003/1, the BBSRC grant BB/S006338/1, and by the EU TAILOR grant. We also acknowledge the use of the EPSRC-funded Tier 2 facility
JADE (EP/P020275/1) and GPU computing support by Scan Computers International Ltd.

\bibliographystyle{unsrtnat}
\bibliography{references}  

\appendix

\section{Details on Evaluation Metrics}

In this section, we explain the metrics used in the main body of this work in more detail. We employ scalar quantification metrics, such as expected calibration error (ECE) and maximum calibration error, together with visual tools, such as reliability diagrams and confidence distribution histograms \citep{guo2017calibration} and classification margin diagrams \citep{dai2018adversarial} to evaluate model calibration.

The term ``confidence calibration'' is used to describe the ability of a model to produce probability estimations that are accurate reflections of the correct likelihood of an event, which is imperative especially in real-world applications. Considering a $K$-class classification task, let  $X \in \mathcal{X}$ and $Y \in \mathcal{Y}=\{1, \ldots, K\}$ be input and true ground-truth label random variables, respectively. Let $\hat{Y}$ denote a class prediction and $\hat{P}$ be its associated confidence, i.e., probability of correctness. We would like the confidence estimate $\hat{P}$ to be calibrated, which intuitively means that $\hat{P}$ represents a true probability. The perfect calibration can be described as follows:
\begin{equation}
\label{eqn:calibration}
    \mathbb{P}(\hat{Y}=Y \mid \hat{P}=p)=p, \quad \forall p \in[0,1]\,.
\end{equation}
\textbf{ECE} captures the notion of average miscalibration in a single number, and can be obtained by the expected difference between  accuracy and confidence of the model: $
\underset{\hat{P}}{\mathbb{E}}[|\mathbb{P}(\hat{Y}=Y \mid \hat{P}=p)-p|]\,.
$

\textbf{MCE} is  crucial in safety- and security-critical settings, as it quantifies the worst-case expected miscalibration: $
\max _{p \in[0,1]}|\mathbb{P}(\hat{Y}=Y \mid \hat{P}=p)-p| .
$

\textbf{Reliability diagrams} are visual representation tools for model calibration, as Equation~\ref{eqn:calibration} is intractable, because  $\hat{P}$ is a continuous random variable. They characterise the average accuracy level inside points from a given confidence level bin.

\textbf{Classification margin} is simply the difference between the model output probability of the ground-truth class and the model probability of the predicted most-likely class, i.e., the probability of the best second class. Thus, this metric is between $-1$ and $1$, where values close to $-1$ indicate that a model is overconfident in wrong predictions, and values closer to $1$ indicate that the model is confident in its correct prediction. In our reported class margin diagrams,  we average these values over many samples and repeated trials with different random seeds to draw box-plot diagrams of our results.

\section{Reproducibility}

We use standard splits on all datasets. We report the average results of 5 runs on different seeds; the hyperparameters are selected using the validation set. Following the results from the original papers on the baseline models, we evaluate our model on 2 GNN layers. The models are trained for 300 epochs on citation graphs and 50 epochs on PPI. We also use the Adam optimiser. The reported results on the calibration analysis were performed with the initial learning rate of 0.001 for both GCNs and GPCNs, as it provided the best performances for both models. For adversarial attacks, we set the initial learning rate to 0.01 and the number of epochs to 200 to compare our results to other works. For the general performance experiment, we use grid search for the hyperparameters, as described in Table~\ref{tab:hyperpar}. Inductive tasks were trained using a GCN version of GraphSage \citep{hamilton2017inductive} with the neighborhood sample size of 25 and 10 for the first and the second GNN layer, respectively (see \citep{hamilton2017inductive} for more details).

Our experiments are performed using the PyTorch Geometric library \citep{fey2019fast}. To do an extensive experiment, we build on GraphGym \citep{you2020design}, a research platform for designing and evaluating GNNs, and we seamlessly integrate it with the predictive coding learning algorithm. In addition, we employ another PyTorch library for adversarial attacks and defenses known as DeepRobust \citep{li2020deeprobust} for various type of adversarial attacks that we perform on graphs.

\subsection{Datasets}
\begin{table}[H]
    \caption{An overview of the data sets used in our experiments }\label{tab:datasets}
    \begin{adjustbox}{max width=\textwidth}
    \begin{tabular}{*{7}{|c}|}\hline
       & CORA & CiteSeer &  PubMed & PPI &  Reddit \\\hline
     \hline
     Type & Citation & Citation & Citation & Protein interaction & Communities  \\
     \hline
     \#Nodes & 2708 & 3327 & 19717 (1 graph) & 56944 (24 graphs) & 232965 (1 graph) \\
     \hline
     \#Edges  & 5429 & 4732 & 44338 & 818716 & 114615892   \\
     \hline
     \#Features/Nodes & 1433 & 3703 & 500 & 50 & 602   \\
     \hline
     \#Classes & 7 & 6 & 3 & 121(multilabel) & 41   \\
     \hline
     \#Training Nodes & 140 & 120 & 60 & 44906(20 graphs) & 153431    \\
     \hline
     \#Validation Nodes & 500 & 500 & 500 & 6513 (2 graphs) & 23831  \\
     \hline
     \#Testing Nodes  & 1000 & 1000 & 1000 & 5524 (2 graphs) & 55703   \\
     \hline
 \end{tabular}
 \end{adjustbox}
\end{table}

\begin{table}[H]
  \centering
  \caption{Hyper-Parameter Search}\label{tab:hyperpar}  
  \begin{tabular}{|c|c|c|}
    \hline
    Parameter Type & Grid \\
    \hline
    values nodes update rate & $0.05, 0.1, 0.5, 1.0$  \\
    Weight update learning rate & $1e-2$,$1e-3$,$1e-4$,$1e-5$ \\
    Number of GNN Layer & $2, 4$ \\
   Inference steps,T, & 12, 32, 50, 100\\
    PC synaptic weight update rate &  at the end of ,T, inference steps, and at every inference step\\
    aggregation functions  &  sum, add, max\\
    Graphsage sampling & 10, 25 for first and second GNN layer respectively\\
    \hline
  \end{tabular}
  
\end{table}

\section{Calibration Analysis: Confidence in Prediction} 
\begin{figure}[H]
    \centering
    {{\includegraphics[width=6.5cm]{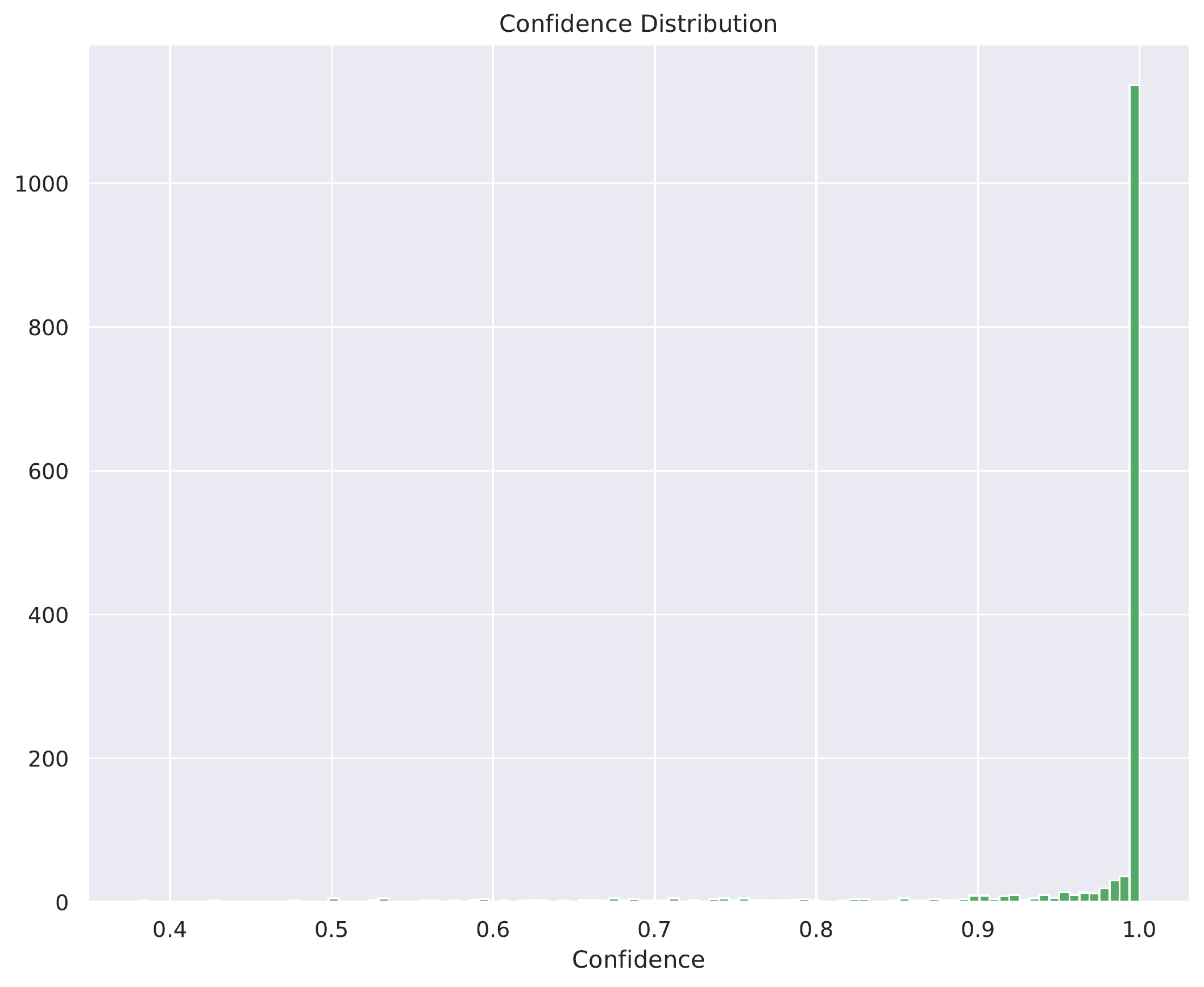} }}%
    \qquad
    {{\includegraphics[width=6.5cm]{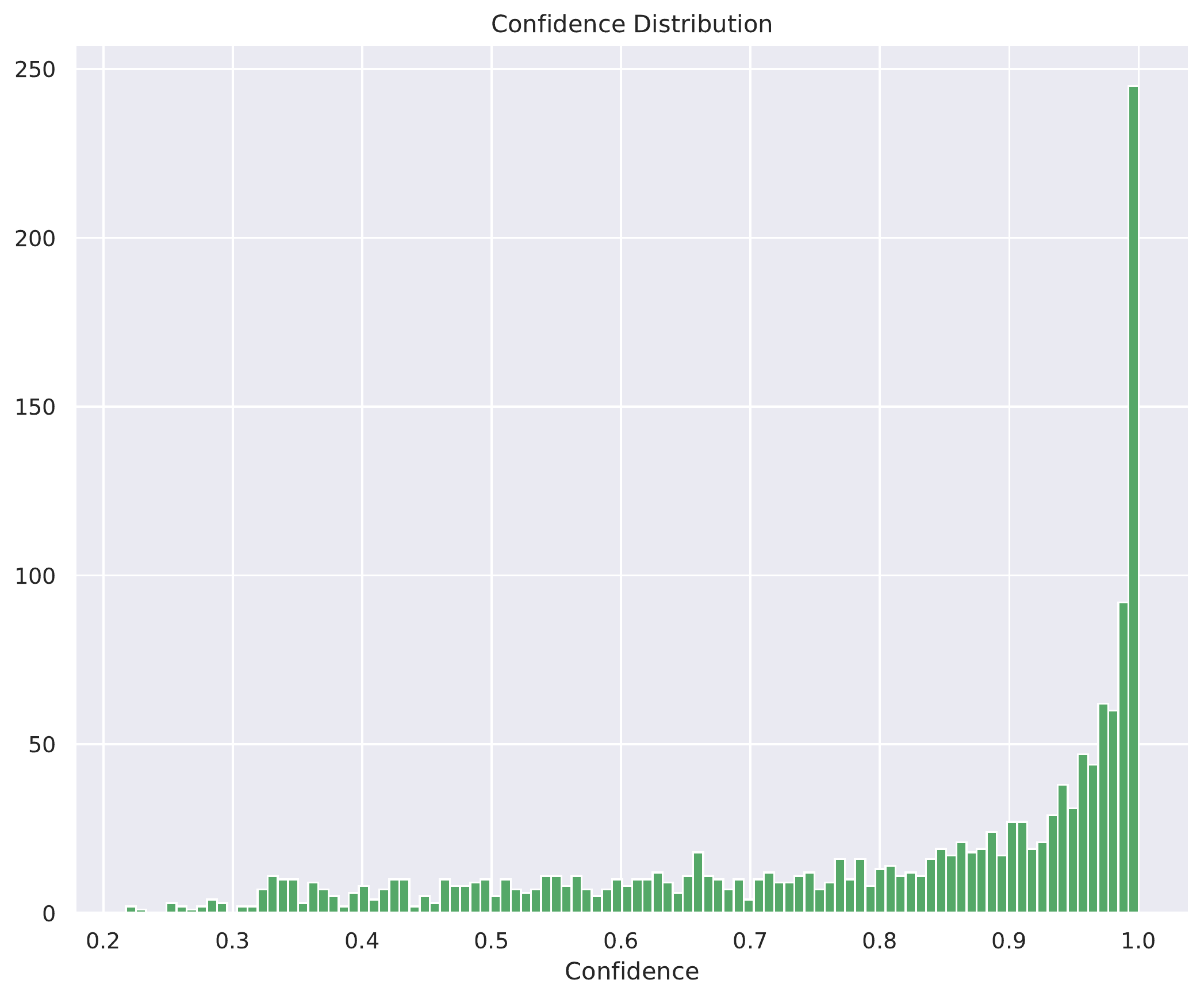} }}%
    \caption{Histogram of the prediction confidence distribution on the CORA dataset. The x-axis indicates the confidence of the model on the samples, and the y-axis is the count on a normal scale of data points that fall into a given confidence bin. Left: GCN. Right: GPCN.}%
    \label{fig:conf_distribution_cora}%
\end{figure}

\begin{figure}[H]
    \centering
    {{\includegraphics[width=6.5cm]{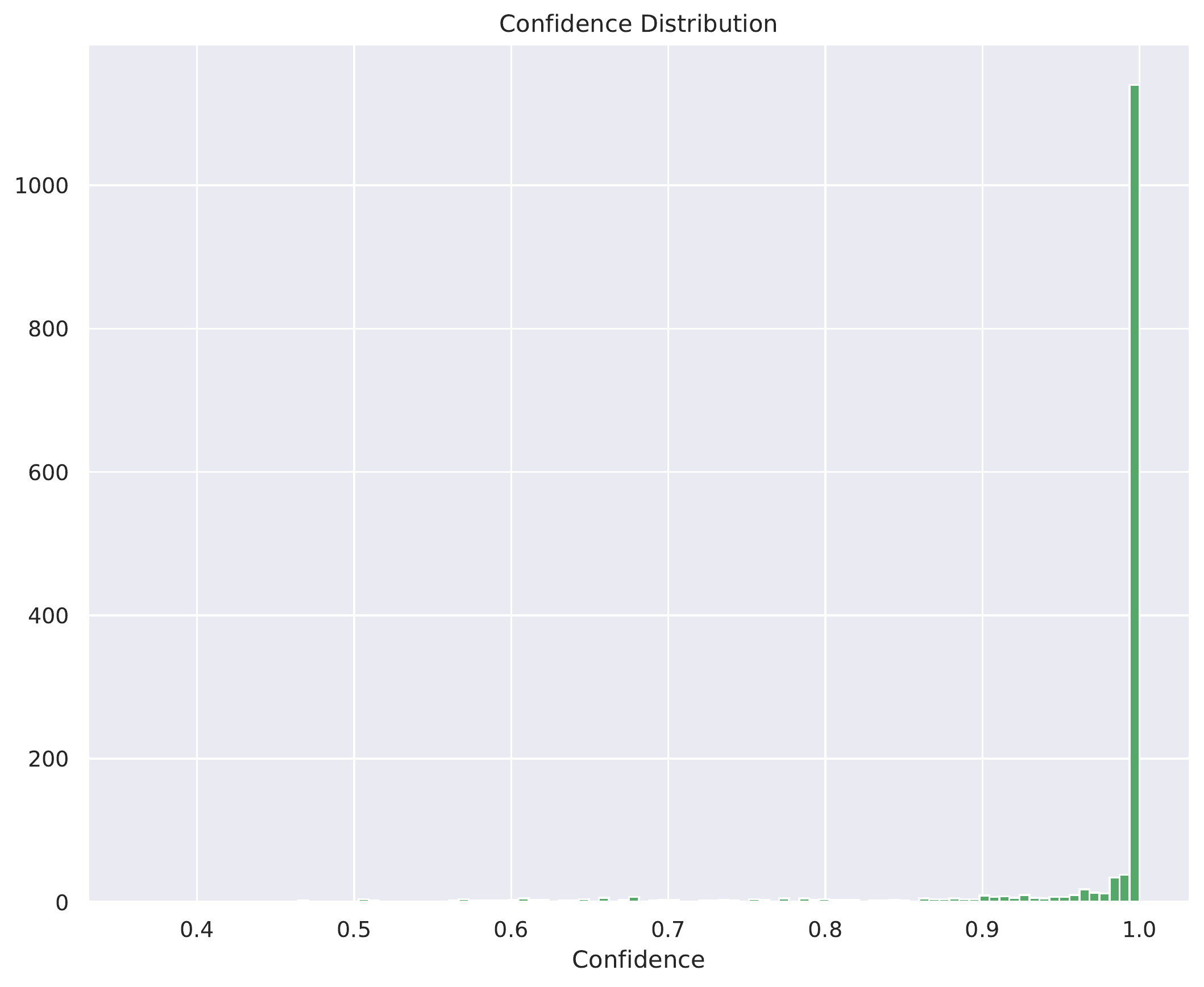} }}%
    \qquad
    {{\includegraphics[width=6.5cm]{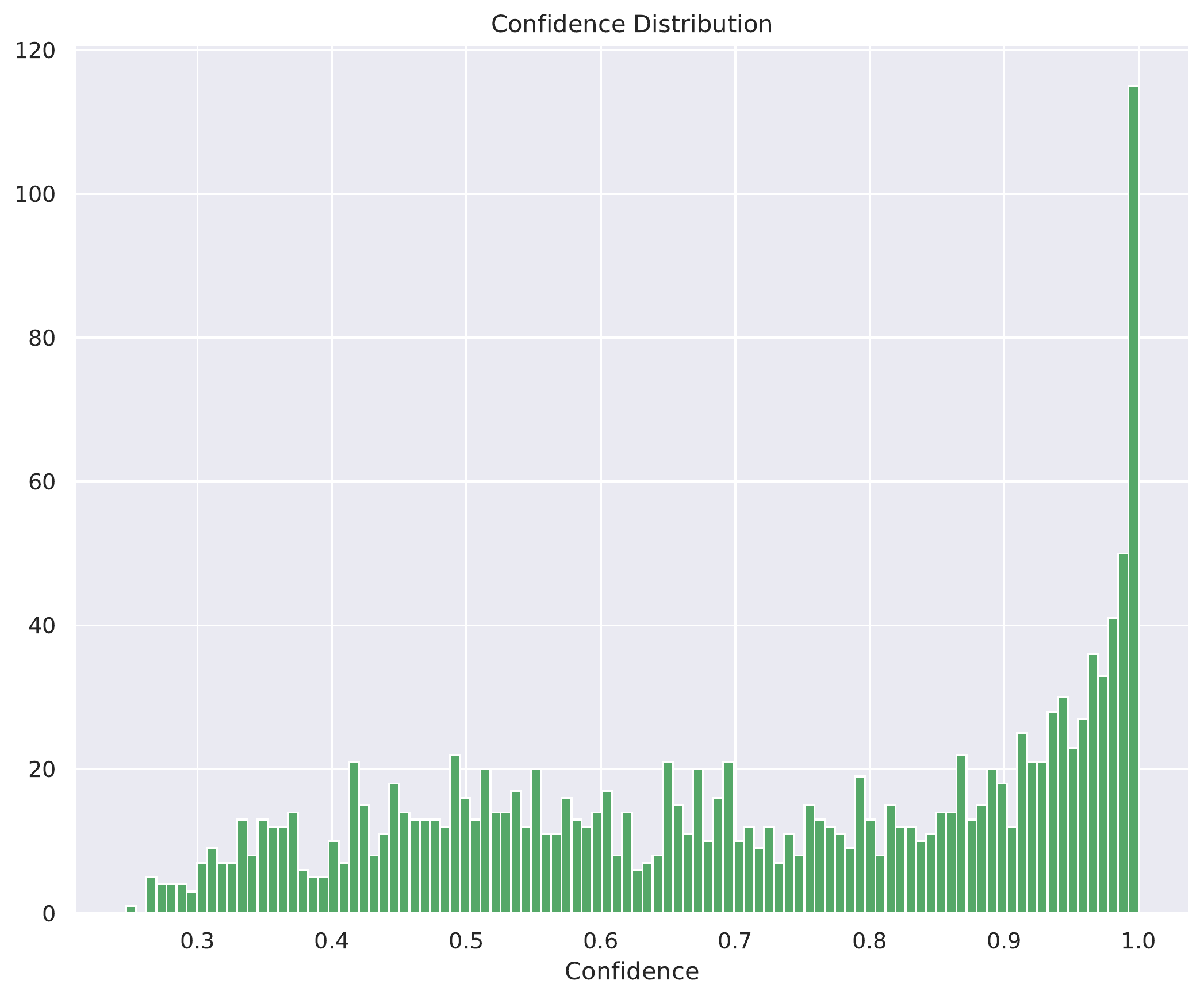} }}%
    \caption{Histogram of the prediction confidence distribution on the CiteSeer dataset. The x-axis indicates the confidence of the model on the samples, and the y-axis is the count on a normal scale of data points that fall into a given confidence bin. Left diagram is for GCN while right diagram is ouput of GPCN.}%
    \label{fig:conf_distribution_citeseer}%
\end{figure}

\begin{figure}[H]
    \centering
    {{\includegraphics[width=6.5cm]{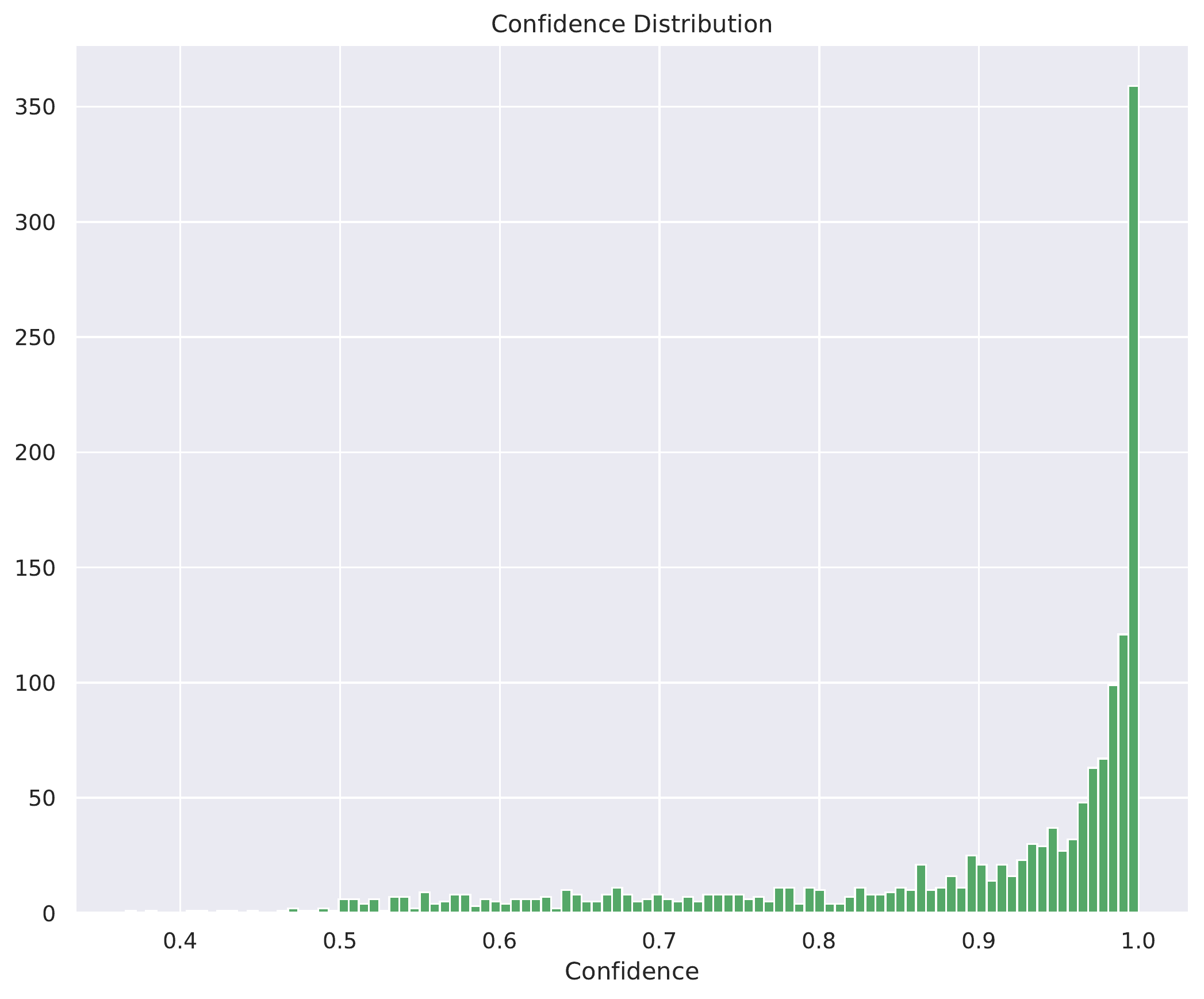} }}%
    \qquad
    {{\includegraphics[width=6.5cm]{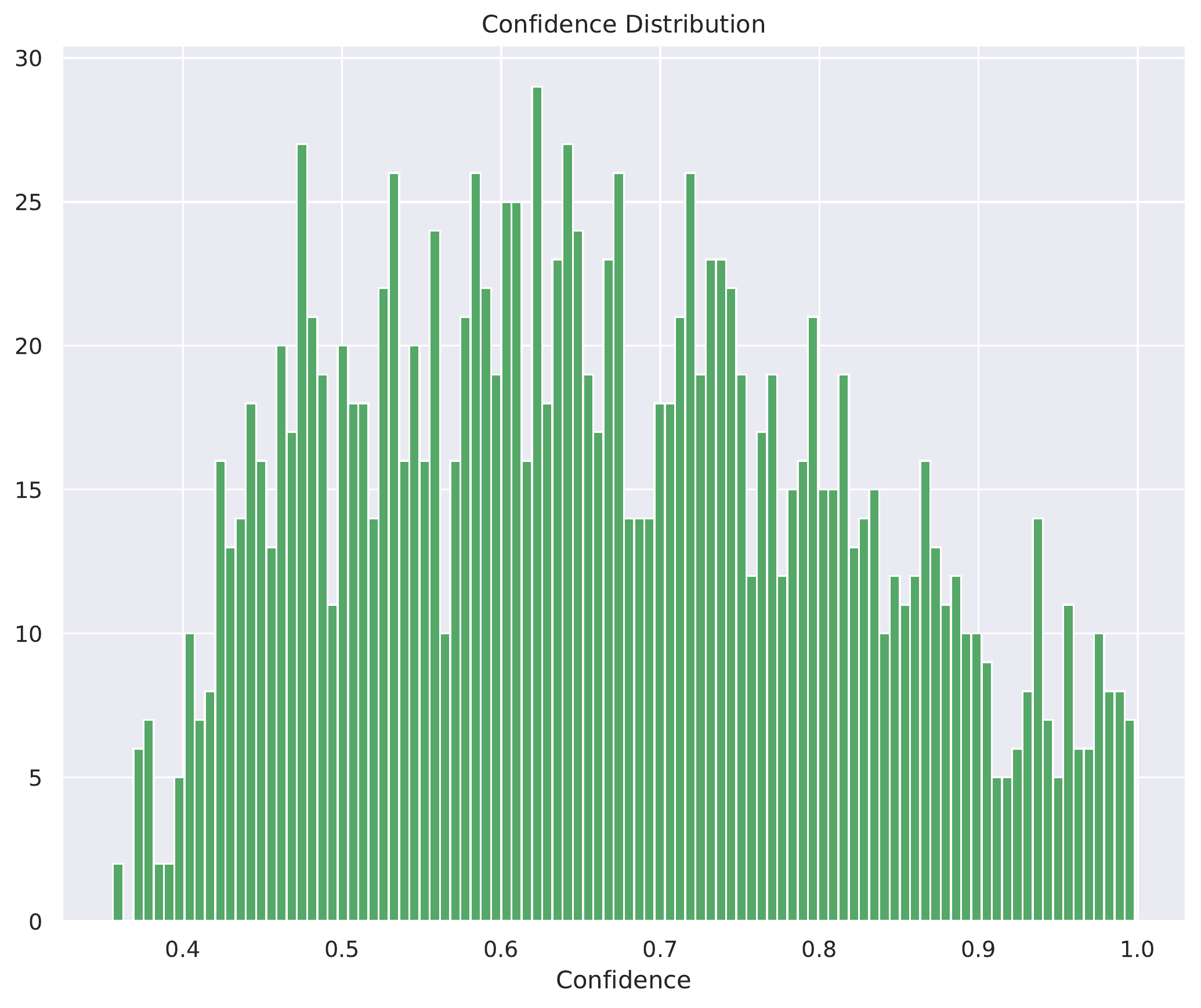} }}%
    \caption{Histogram of the prediction confidence distribution on the PubMed dataset. The x-axis indicates the confidence of the model on the samples, and the y-axis is the count on a normal scale of data points that fall into a given confidence bin. Left: GCN. Right: GPCN.}%
    \label{fig:conf_distribution}%
\end{figure}

As deep networks tend to be overconfident even if they are wrong, we compared the confidence distribution of GPCNs with GCNs in Figs~\ref{fig:conf_distribution_cora},~\ref{fig:conf_distribution_citeseer}, and \ref{fig:conf_distribution}, where confidence is the maximum of the softmax of the model output. The result in this section correspond to the reported results in the body of the paper on calibration analysis in Section~\ref{sec:cal_analysis}. Both models are run for 300 epochs using the same parameters (i.e., learning rate on weights equal to 0.001), and we select the best model based on the validation set for GCNs. For GPCNs, we select the best model based on the best accuracy on the validation set as well as the lowest energy on the training set, as the energy minimization can be interpreted as likelihood maximisation. We consider the energy while selecting the best model for evaluation, because we discovered a high correlation between energy and robustness, as we will demonstrate in the following section (see Fig.~\ref{fig:energy_calibration}). Interestingly, the results on the prediction distribution provide another dimension to communicate the same results that we witness in Section~\ref{sec:cal_analysis} using reliability diagrams. We see that on the prediction distribution on the CORA dataset in Fig.~\ref{fig:conf_distribution}, GPCNs are relatively less confident in their predictions, while GCNs are overly confident. Fig.~\ref{fig:conf_distribution_citeseer} on the CiteSeer dataset similarly shows that most prediction confidences of the GCN model are less than 0.5, showing that GCNs are overly under-confident as we saw in the body of the paper. GPCNs, on the other hand, provide a well-behaved prediction confidence distribution on Citeseer: demonstrating that GCNs are either under-confident despite a high-performance accuracy or over-confident in the manner that is disproportional to the  performance accuracy.

\subsection{Calibration Strengthening through Energy Minimization }

\begin{figure}[H]
    \centering
    \makebox[\textwidth]{
    \includegraphics[width=12cm]{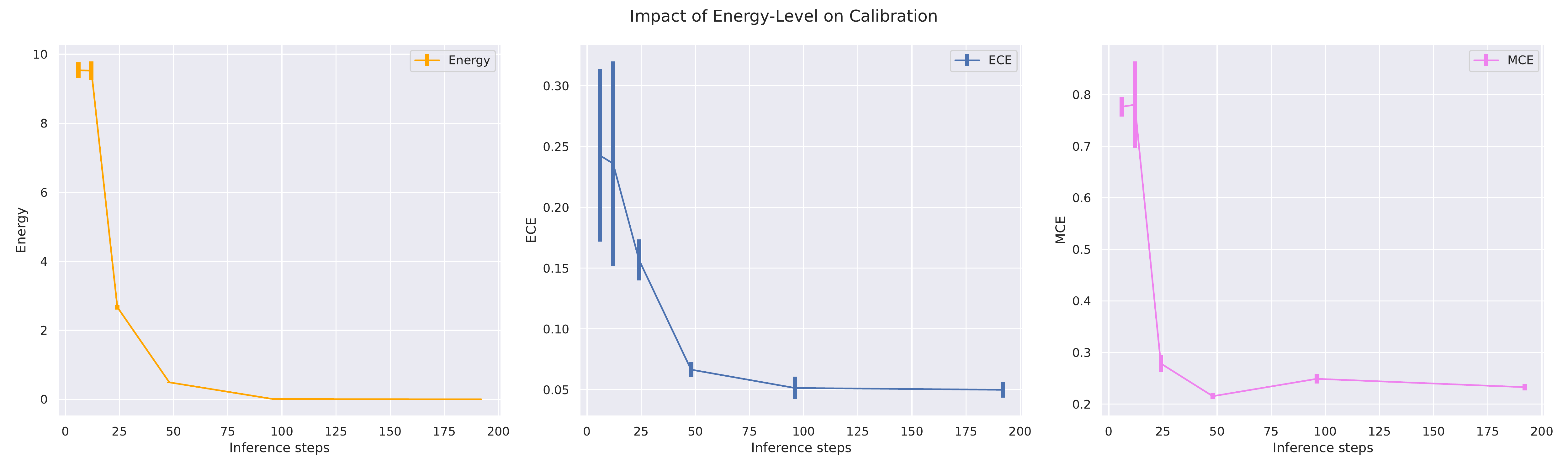}
    }
    \caption{Evaluation of the impact of the energy on the calibration performance on CORA  (learning rate $= 0.001$). Left: Demonstrates that increasing inference steps leads to a lower energy. Middle: Shows that the lower energy level determines the expected error (ECE). Right: Also showcases  the correlation between the energy and the maximum calibration error (MCE).}
    \label{fig:energy_calibration}
\end{figure}

\begin{figure}[H]
    \centering
    \makebox[\textwidth]{
    \includegraphics[width=12cm]{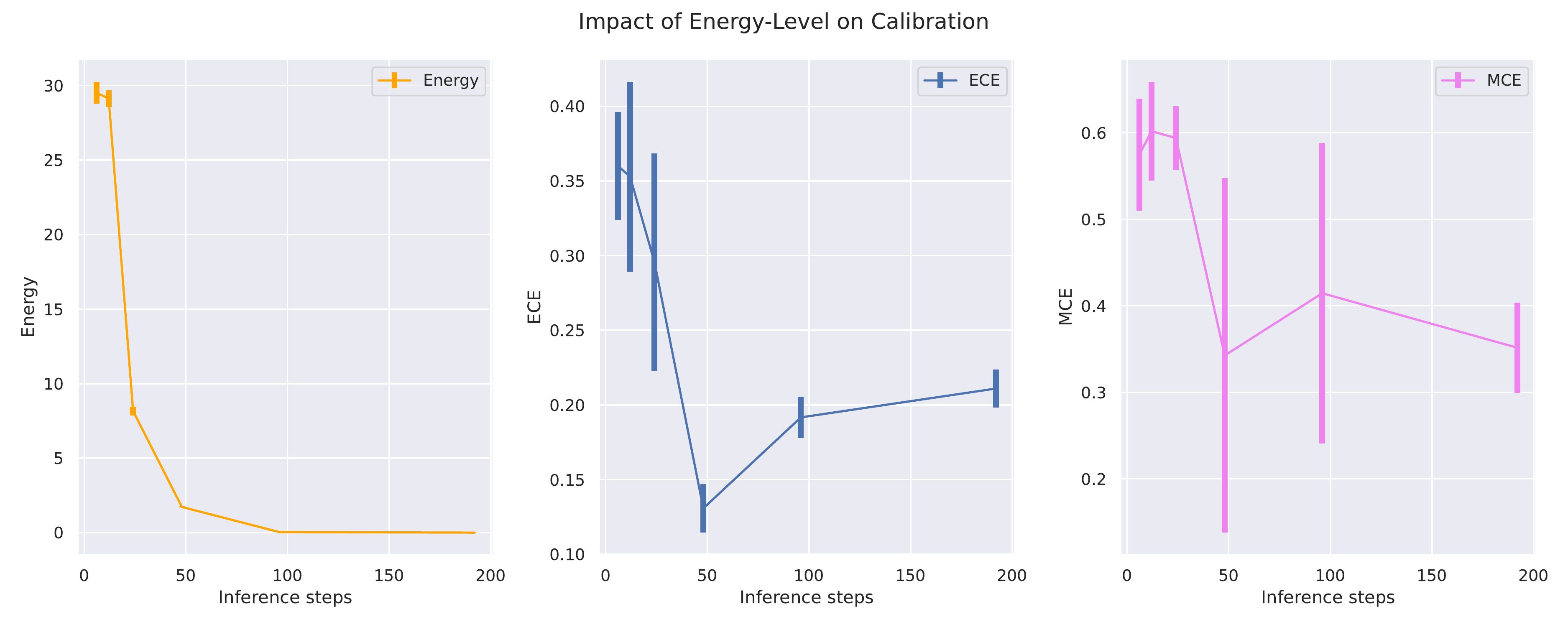}
    }
    \caption{Evaluation of the impact of the energy on the calibration performance on CiteSeer. Left: Demonstrates that increasing inference steps leads to a lower energy. Middle: Shows that the lower energy level determines the expected error (ECE). Right: Also showcases the correlation between the energy and the maximum calibration error (MCE).}
    \label{fig:energy_calibration_citeseer}
\end{figure}

We observed a high positive correlation between calibration and training energy level, thus we further investigated the role of the energy to the robustness and calibration of our learned representation. Differently from standard predictive coding networks, we observed that GPCNs require several inference steps to reach the lowest training energy possible, i.e., this can seen as reaching a local optimum of the likelihood maximisation function. More importantly, we also observed that the lower the energy the better the calibration is, i.e., the better the model can estimate uncertainty in its prediction. Figure~\ref{fig:energy_calibration} 
on the CORA dataset and Fig.~\ref{fig:energy_calibration_citeseer} on the CiteSeer dataset showcase this correlation. The plots on the left show that the inference steps correlate to the energy level, i.e., the longer the inference is, the more likely that the model converges to a lower energy. The middle diagram and right diagrams, similarly, present the correlation between the inference steps and ECE and MCE, respectively, which from the left diagrams, implies the correlation of the energy level and ECE and MCE. We see that the lower the energy the better the calibration performance reached.

\begin{figure}
    \centering
{{\includegraphics[width=6.5cm]{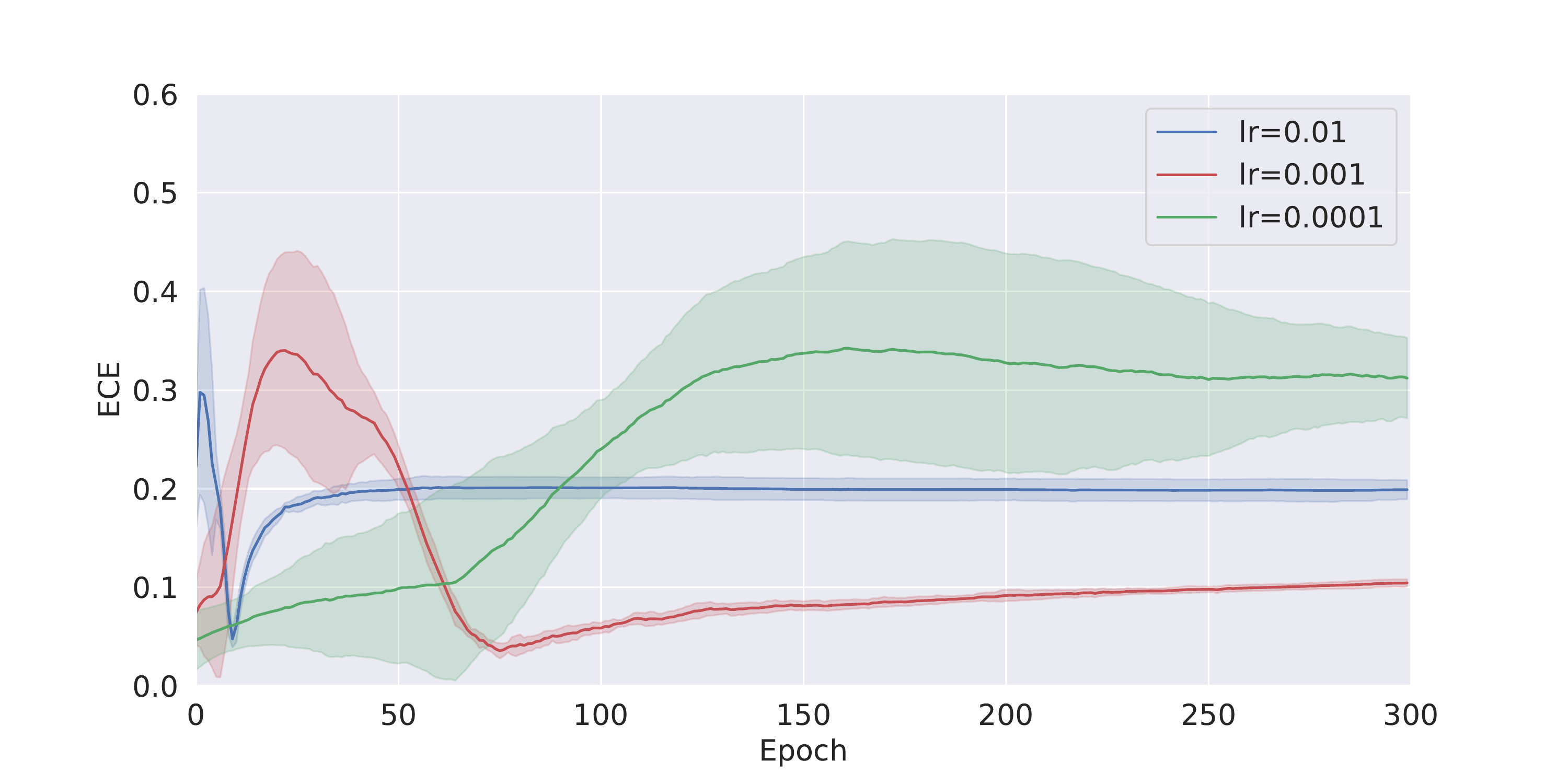} }}%
    \qquad
    {{\includegraphics[width=6.5cm]{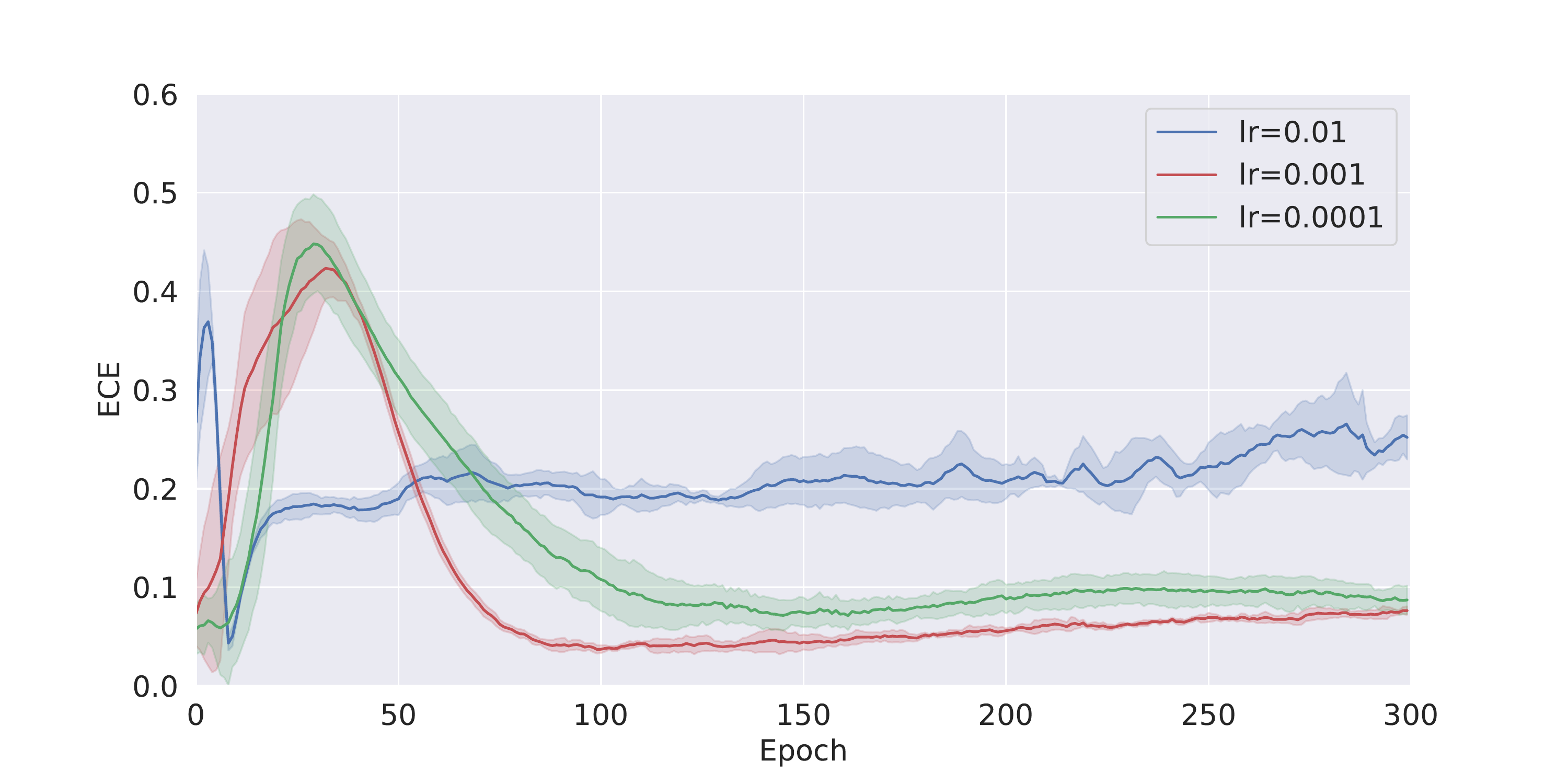} }}%
    \caption{Evolution of expected calibration error (ECE) on the test set during training on various learning rates (lr). Left: GCN. Right: GPCN.}%
    \label{fig:ece_lr}%
\end{figure}

\begin{figure}
    \centering
{{\includegraphics[width=6.5cm]{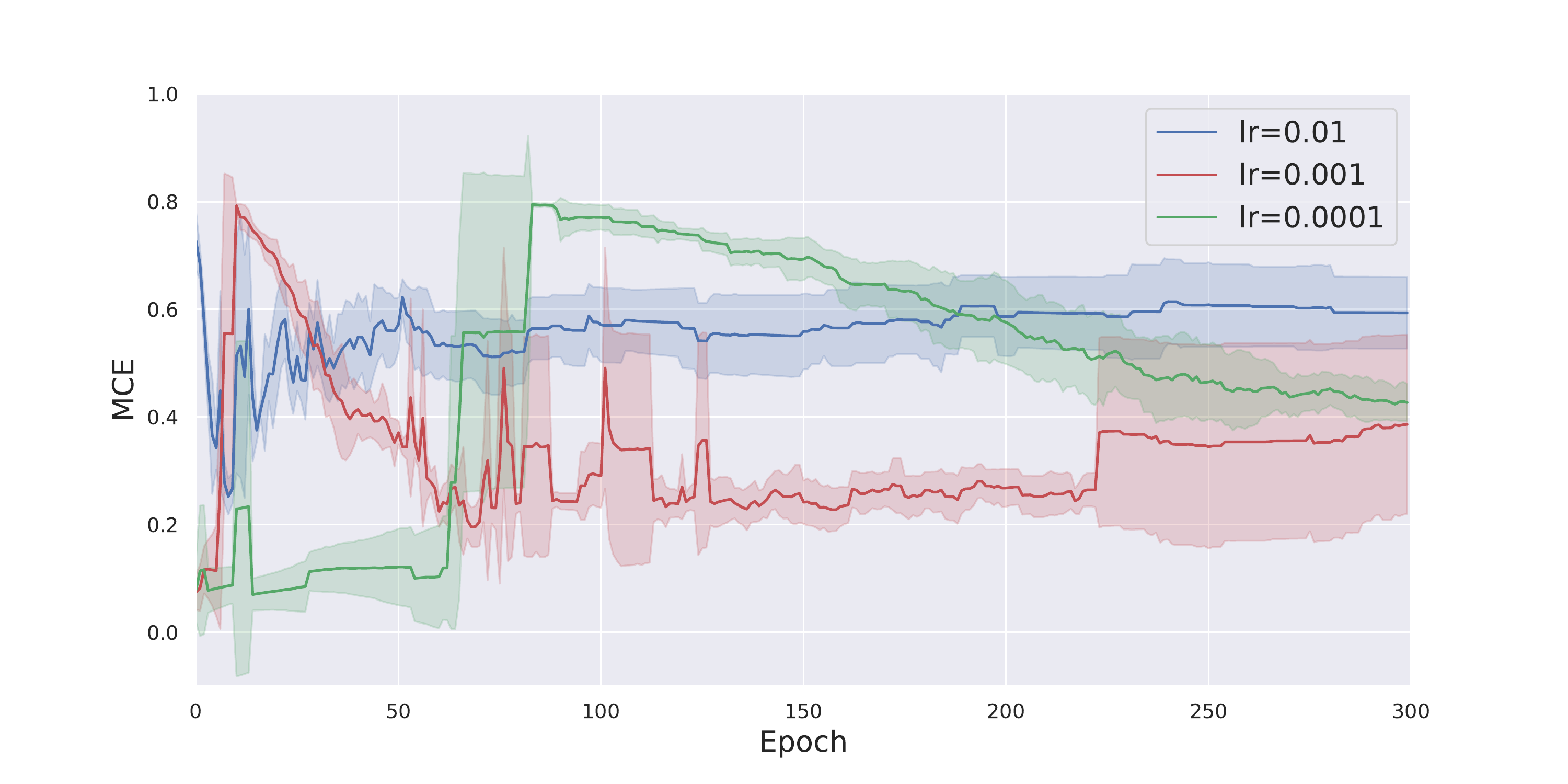} }}%
    \qquad
{{\includegraphics[width=6.5cm]{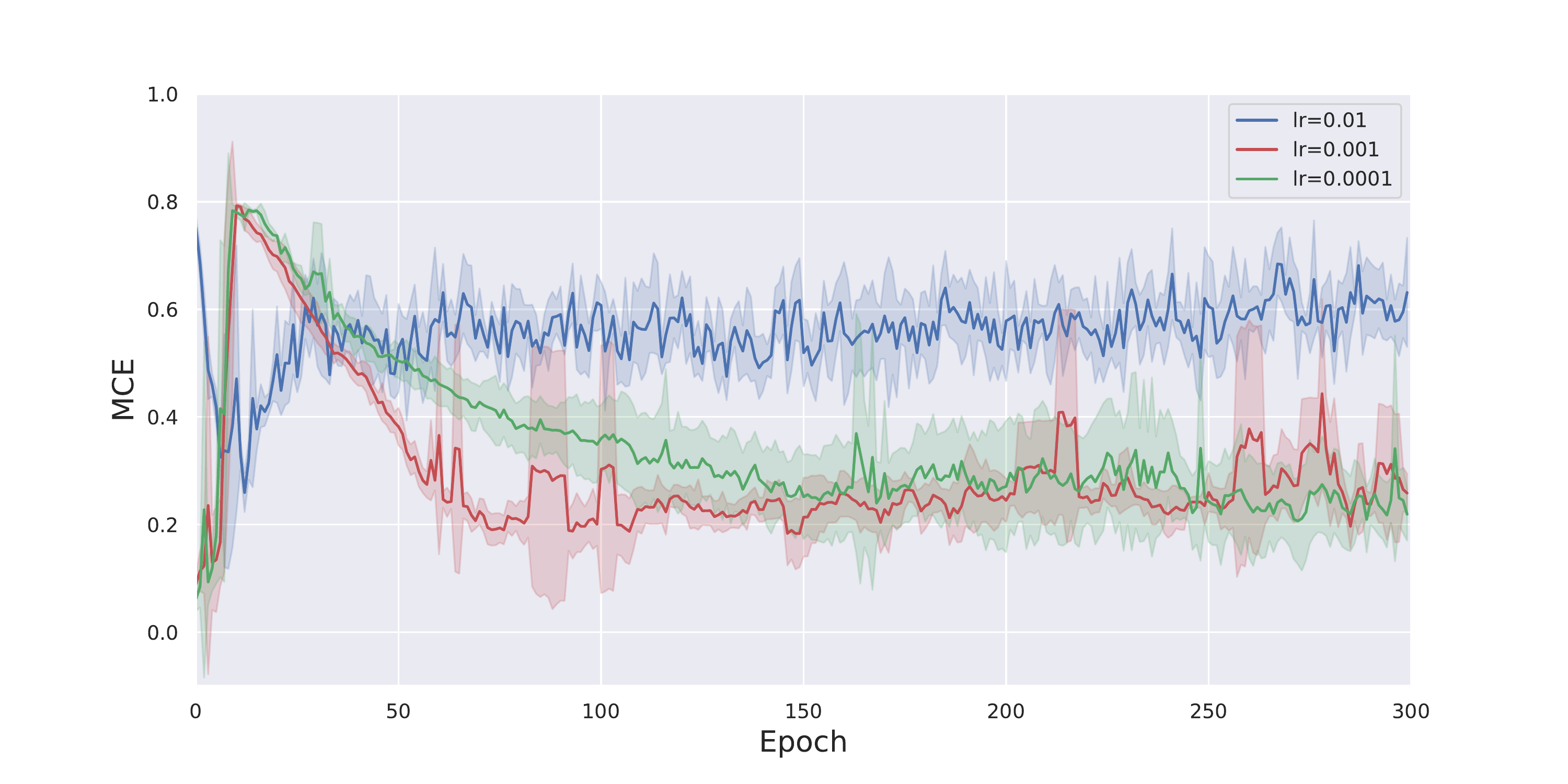} }}%
    \caption{Evolution of maximum calibration error (MCE) on the test set during training on various learning rates (lr). Left: GCN. Right: GPCN.}%
    \label{fig:mce_lr}%
\end{figure}

As it has been shown that calibration can be affected by learning rates \citep{guo2017calibration}, we track the ECE and MCE throughout training, and we see that the lower the learning rate, the better and more stable calibration GPCN is able to attain based on the ECE and MCE metrics (see Figs. ~\ref{fig:ece_lr} and \ref{fig:mce_lr}).

\section{GPCN Architecture}

\begin{figure}[H]
    \centering
    \includegraphics[width=3cm]{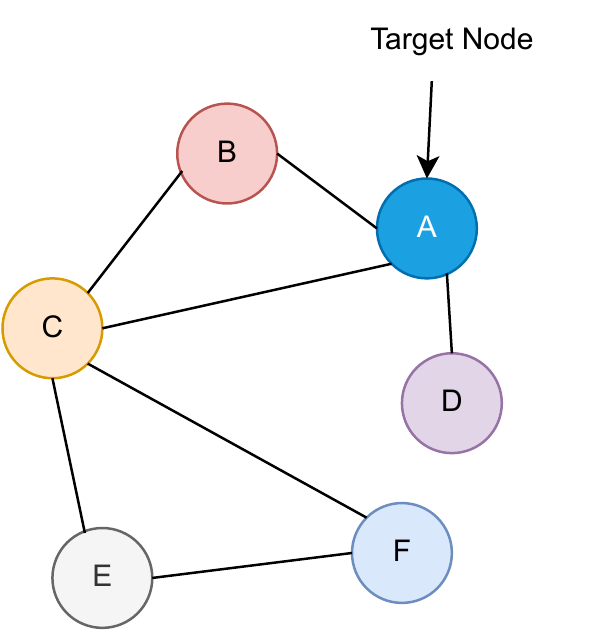}
    \caption{Example of a graph that we use to illustrate the architecture.}
    \label{fig:example_graph}
\end{figure}

\begin{figure}[H]
    \centering
    \makebox[\textwidth]{
    \includegraphics[width=\textwidth]{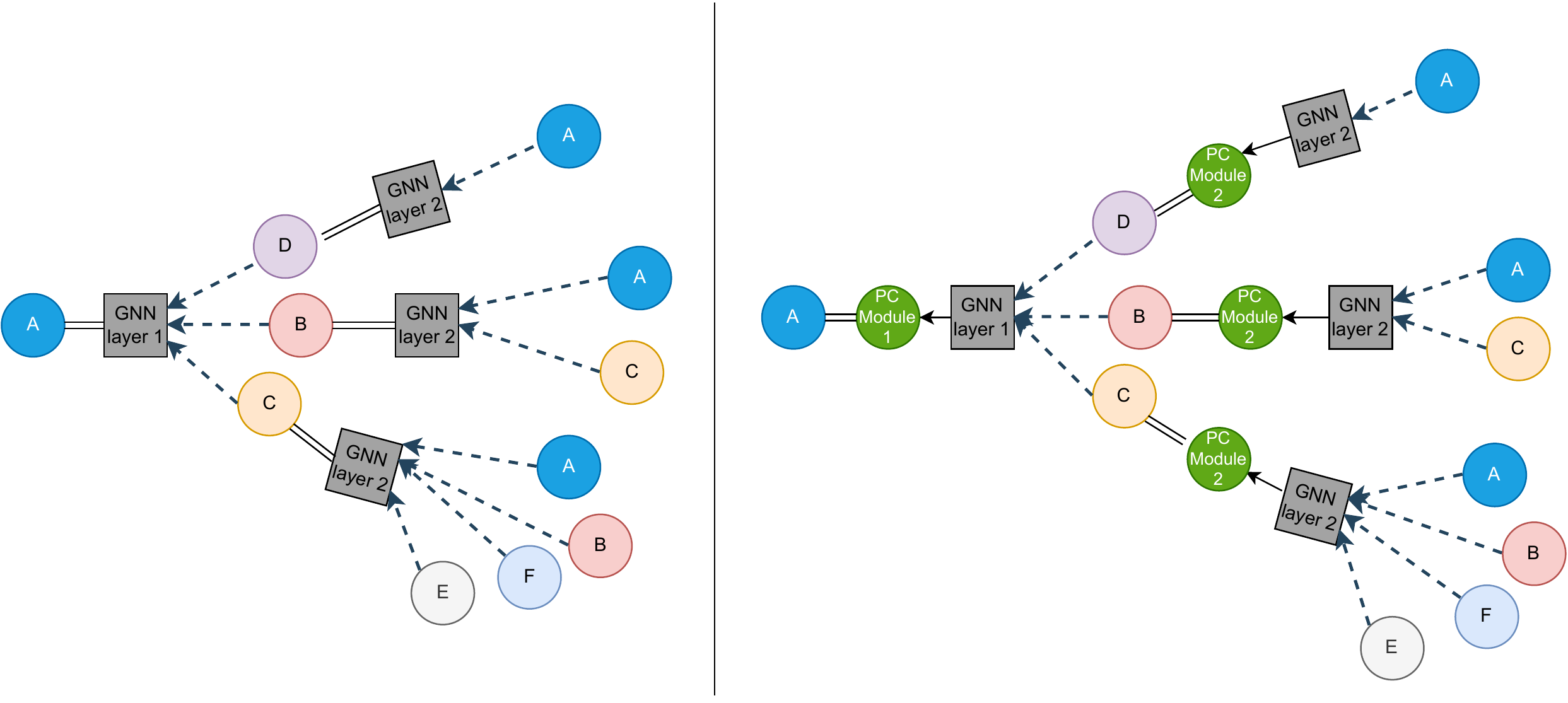}
    }
    \caption{Distinction of messaging propagation between standard GNNs (left) and inter-layer GPCNs (right).}
    \label{fig:graph_mesage_passing}
\end{figure}

Using a simple graph in Fig.~\ref{fig:example_graph}, here, we demonstrate the idea of graph predictive coding as opposed to standard graph neural networks. In typical graph convolution networks (GCNs) \citep{welling2016semi}, the node representation is obtained by the recursive aggregation of representations of its neighbours, after which a learned linear transformation and non-linear activation function are applied. After the $k_th$ round of aggregation (with $k$ denoting the number of the GCN layer), the representation of a node reflects the underlying structure of its nearest neighbours within $k$ hops. Note that the GCN is one of the simplest GNN models, as the update function equates to only neighbourhood aggregation, that is why in Fig.~\ref{fig:graph_mesage_passing} (left), we only depict the aggregate function, as it captures the update function altogether. 

\begin{figure}[H]
    \centering
    \makebox[\textwidth]{
    \includegraphics[width=\textwidth]{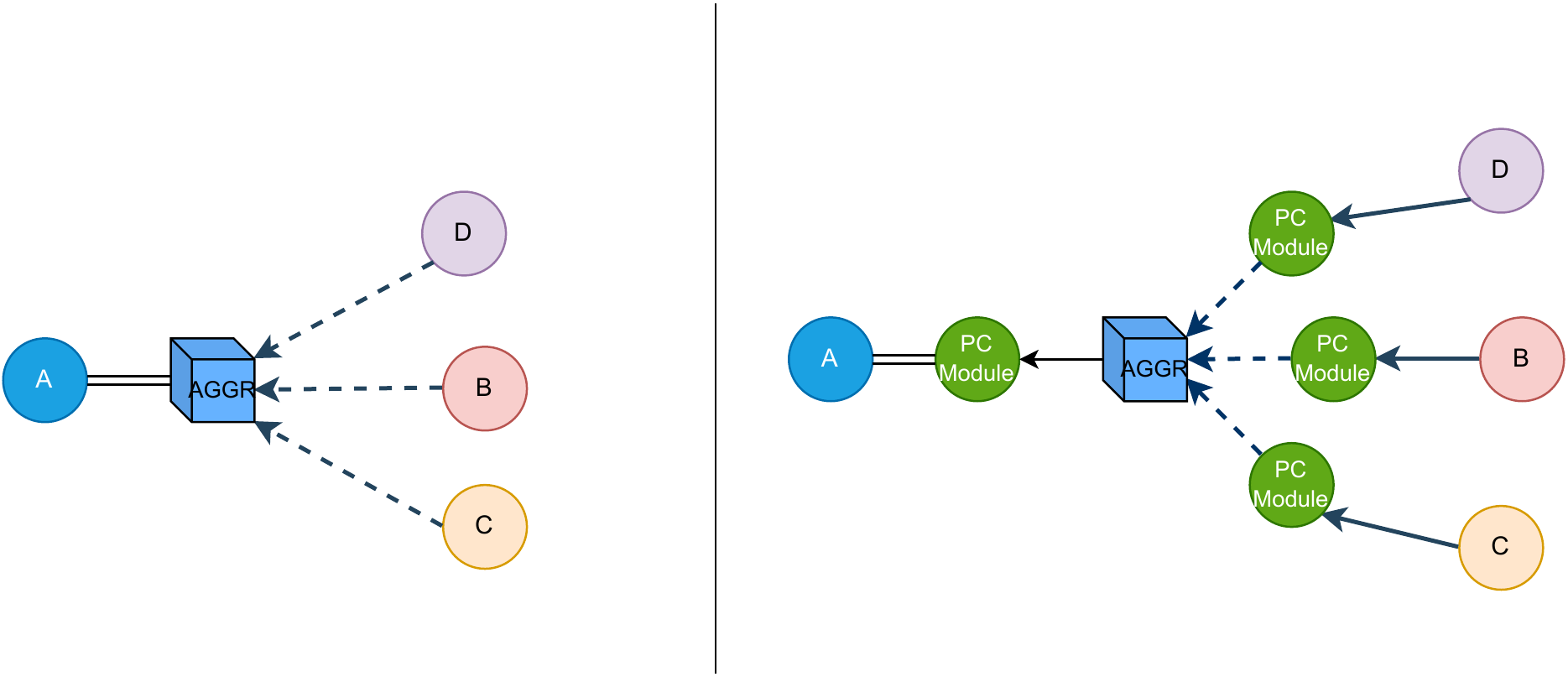}
    }
    \caption{Difference between message aggregation between standard GNNs and GPCNs. Left: standard aggregation method; right: intra-layer GPCNs.}
    \label{fig:node_aggregation}
\end{figure}

Our GPCN model (see Fig.~\ref{fig:graph_mesage_passing} (right)) differs from the standard GNN in three aspects. First, node representation are not a mere result of neighborhood aggregation. Rather, each node has a unique neural state that is updated through energy minimization using the theory of predictive coding described in the main body of this work. Specifically, each neighborhood aggregation at each hop, k, passes through a predictive coding module that predicts the incoming aggregated neighborhood representation. Second, GPCNs have  a different concept of what neighborhood messages are (see Fig.~\ref{fig:node_aggregation}). Rather than transmitting raw messages, the instead forward the residual error of the difference between the predicted representation and the aggregation, which reduces the dynamic range of the message being transmitted, hence acting as a low pass filter. Lastly, unlike standard GNNs that are trained using BP, where the update of weights corresponding to a given neighborhood are dependent, which creates large computation graphs, GPCN learning rules are local and the model weight are updated through energy minimization, as we described in the methodology section.

\section{Additional Results on Evasion Attacks with Nettack}

The following plots demonstrate how both GCNs and GPCNs  perform under various perturbation budgets on the four types of attacks, namely, feature, structure, feature-structure, and indirect attacks. 

\begin{figure}[H]
    \centering
    \makebox[\textwidth]{
    \includegraphics[width=\textwidth]{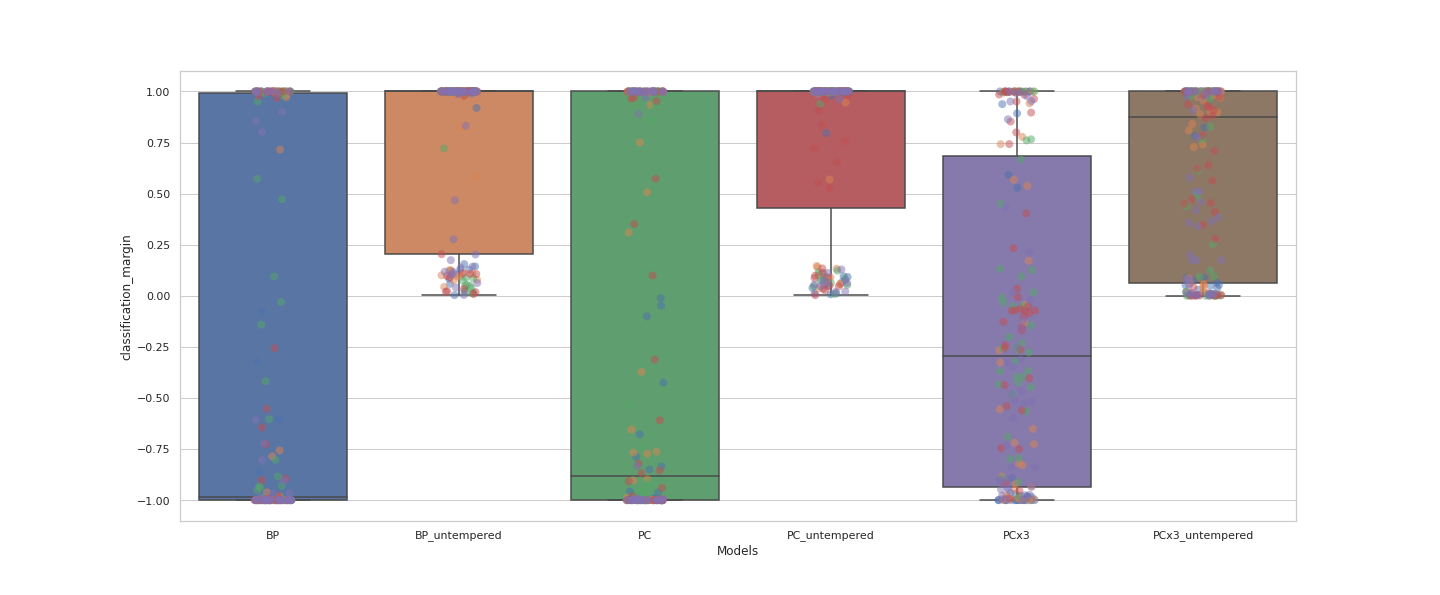}
    }
    \caption{Classification margin diagram of targeted attack on both features and structure when the perturbation rate is 2. On the x-axis, BP indicates the GCN model, PC denotes the GPCN model trained using 12 inference steps, and  PCx3 indicate the GPCN model trained using 36 inference steps, hence achieving a lower training energy. }
    \label{fig:both_s_f_attack1}
\end{figure}

\begin{figure}[H]
    \centering
    \makebox[\textwidth]{
    \includegraphics[width=\textwidth]{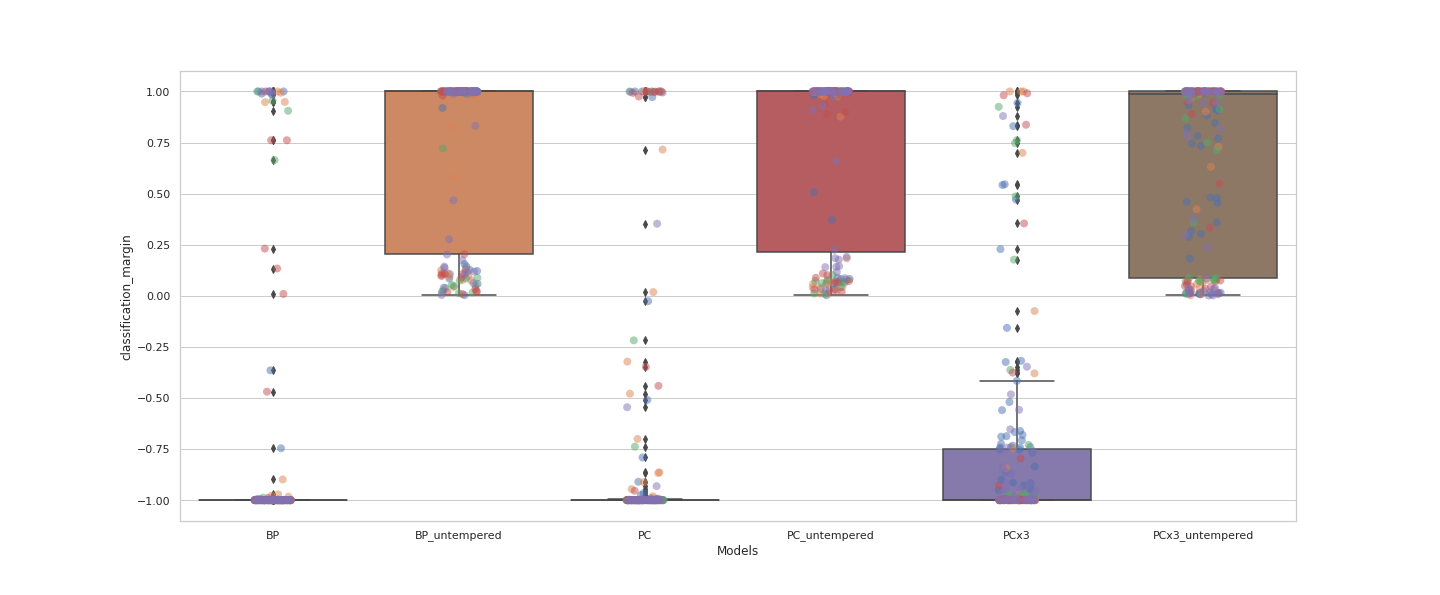}
    }
    \caption{Classification margin diagram of targeted attack on both features and structure when the perturbation rate is~5.}
    \label{fig:both_s_f_attack2}
\end{figure}

\begin{figure}[H]
    \centering
    \makebox[\textwidth]{
    \includegraphics[width=\textwidth]{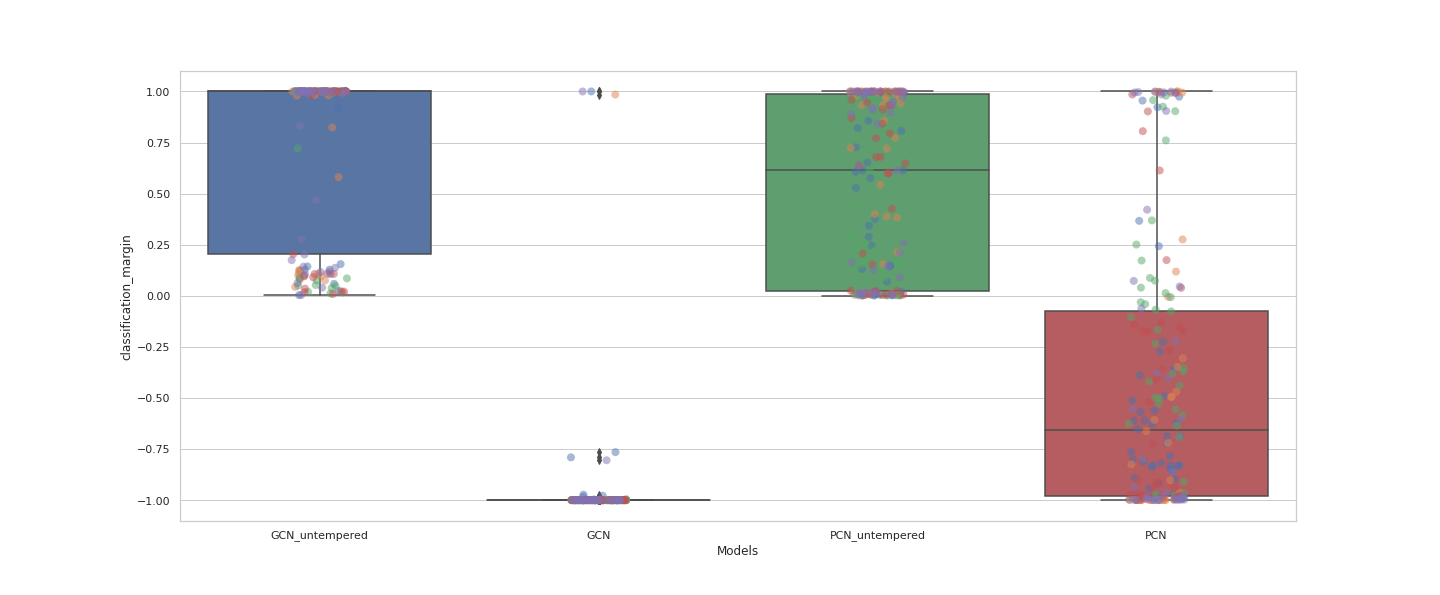}
    }
    \caption{Classification margin diagram of targeted attack on both features and structure when the perturbation rate is~10.}
    \label{fig:both_s_f_attack3}
\end{figure}

\textbf{(1) Structure and feature attack:}
Figure~\ref{fig:both_s_f_attack1} shows that with only 2 perturbations on the neighbourhood structure and features of victim nodes, the median classification margin approaches -1 on the GCN model, while the GPCN model stays relatively robust and with more robustness on a lower energy model (PCx3) where most of the victim nodes have positive classification margins, or in other worlds, they are not adversarially affected by the attack. This trend is even more pronounced when the perturbation rate is increased to 5 (Fig.~\ref{fig:both_s_f_attack2}) and 10 (Fig.~\ref{fig:both_s_f_attack3}), where, except for outliers, the margin of classification of all victim nodes falls to $-1$ for both the GCN and the PC models, and PCx3 stays lately more robust.

\begin{figure}[H]
    \centering
    \makebox[\textwidth]{
    \includegraphics[width=\textwidth]{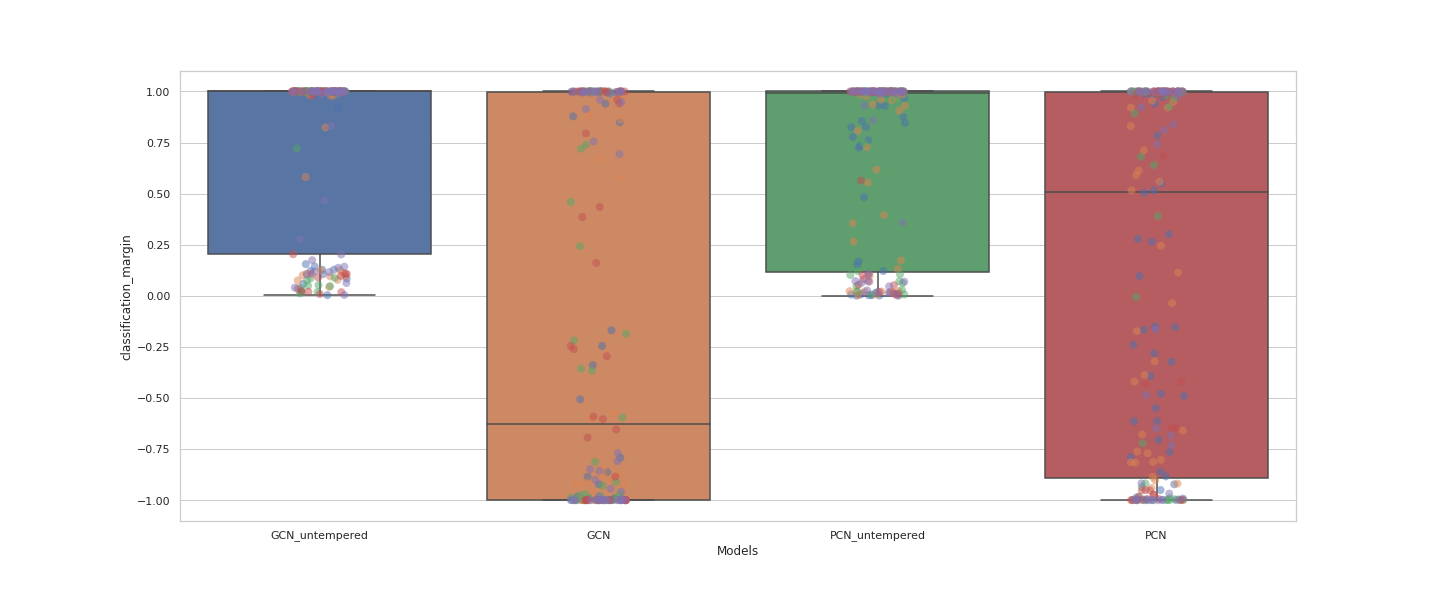}
    }
    \caption{Classification margin diagram of targeted attack on features when the perturbation rate is~30.}
    \label{fig:only_f_attack30}
\end{figure}

\begin{figure}[H]
    \centering
    \makebox[\textwidth]{
    \includegraphics[width=\textwidth]{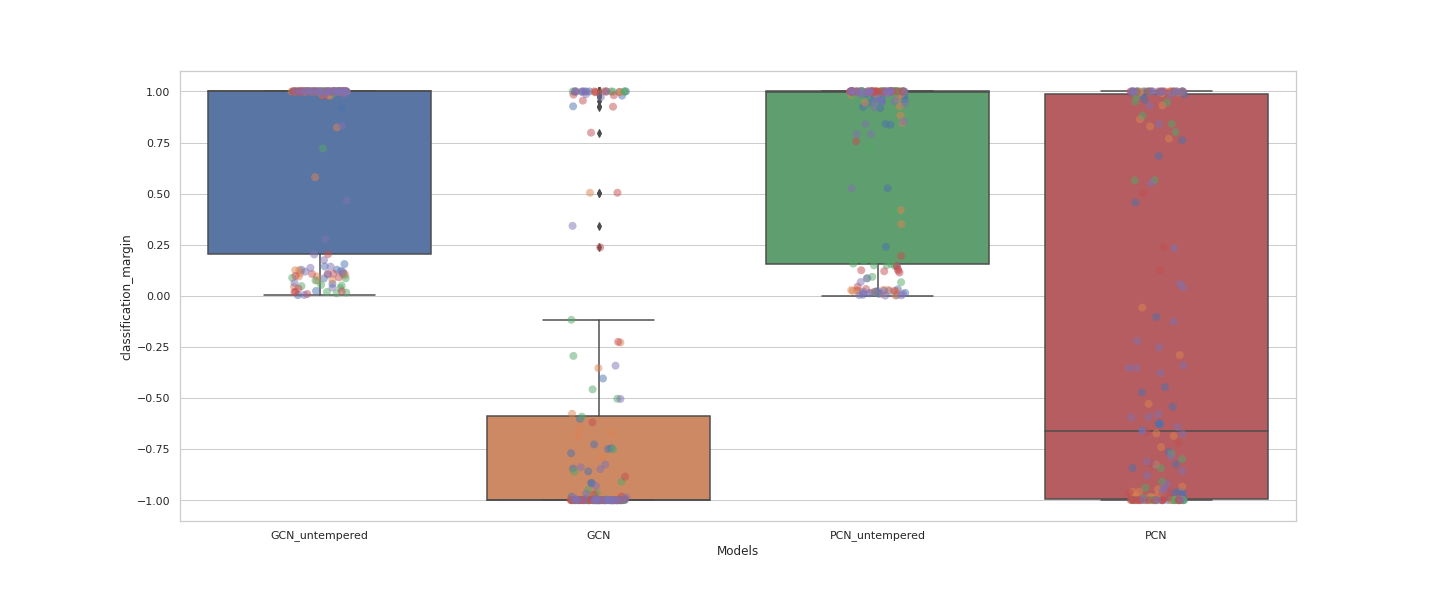}
    }
    \caption{Classification margin diagram of targeted attack on features when the perturbation rate is~100.}
    \label{fig:only_f_attack100}
\end{figure}

\textbf{(2) Feature attacks:}
Since feature attacks do not highly affect GNNs as much as structure attacks, we perform large corruptions of features with the perturbation rate of 1, 5, 10, 30, 50, and 100. We observe similar trends, where small perturbations on features do not affect the model, however, when the perturbation rate becomes  large, our GPCNs display an  unparalleled performance resisting the attacks. When the perturbation rate is  30 (see Fig.~\ref{fig:only_f_attack30}), while the GCN misclassifies around $70\%$ of the victim nodes, the GPCN is still able to classify more than $70\%$ correctly after perturbation. The highly superior performance is observed when the perturbation rate is increased to $100$ in Fig.~\ref{fig:only_f_attack100}, the GCN mislassifies all victim nodes, while the GPCN still classifies correctly those nodes with most victim nodes in the upper quartile having positive classification margins.

\begin{figure}[H]
    \centering
    \makebox[\textwidth]{
    \includegraphics[width=\textwidth]{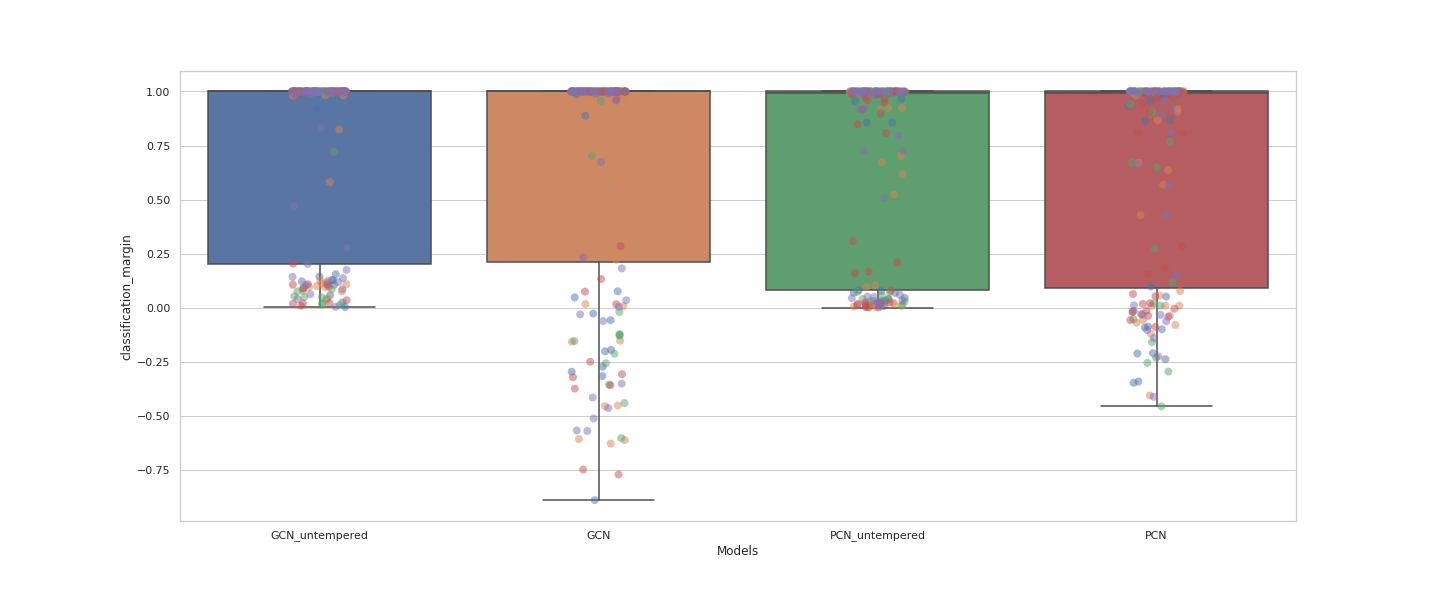}
    }
    \caption{Classification margin diagram of targeted attack on features when the perturbation rate is~1.}
    \label{fig:only_f_attack1}
\end{figure}

\begin{figure}[H]
    \centering
    \makebox[\textwidth]{
    \includegraphics[width=\textwidth]{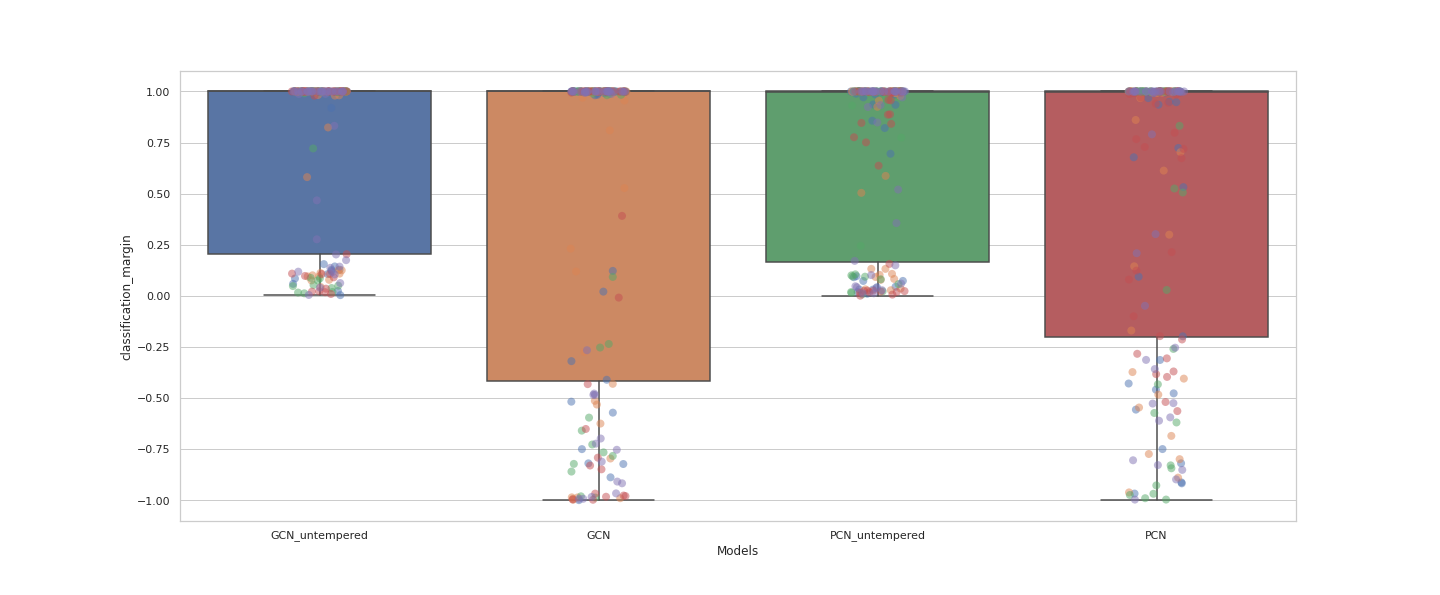}
    }
    \caption{Classification margin diagram of targeted attack on features when the perturbation rate is~5.}
    \label{fig:only_f_attack5}
\end{figure}

\begin{figure}[H]
    \centering
    \makebox[\textwidth]{
    \includegraphics[width=\textwidth]{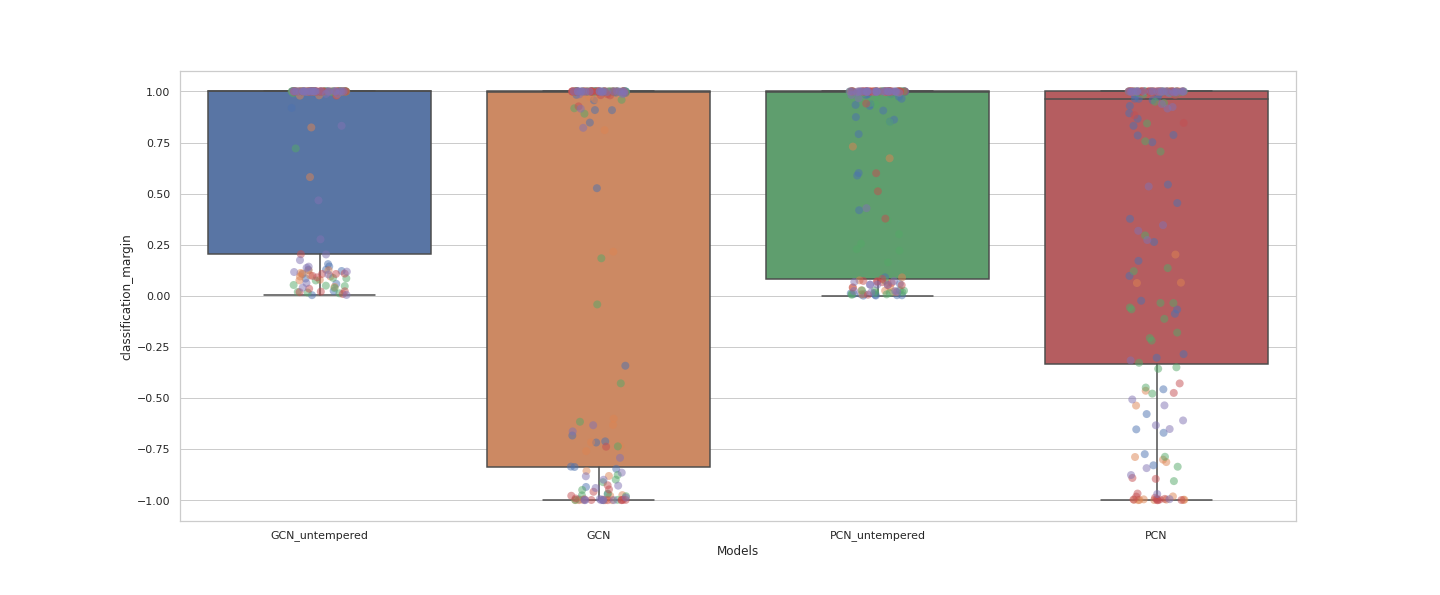}
    }
    \caption{Classification margin diagram of targeted attack on features when the perturbation rate is~10.}
    \label{fig:only_f_attack10}
\end{figure}

\begin{figure}[H]
    \centering
    \makebox[\textwidth]{
    \includegraphics[width=\textwidth]{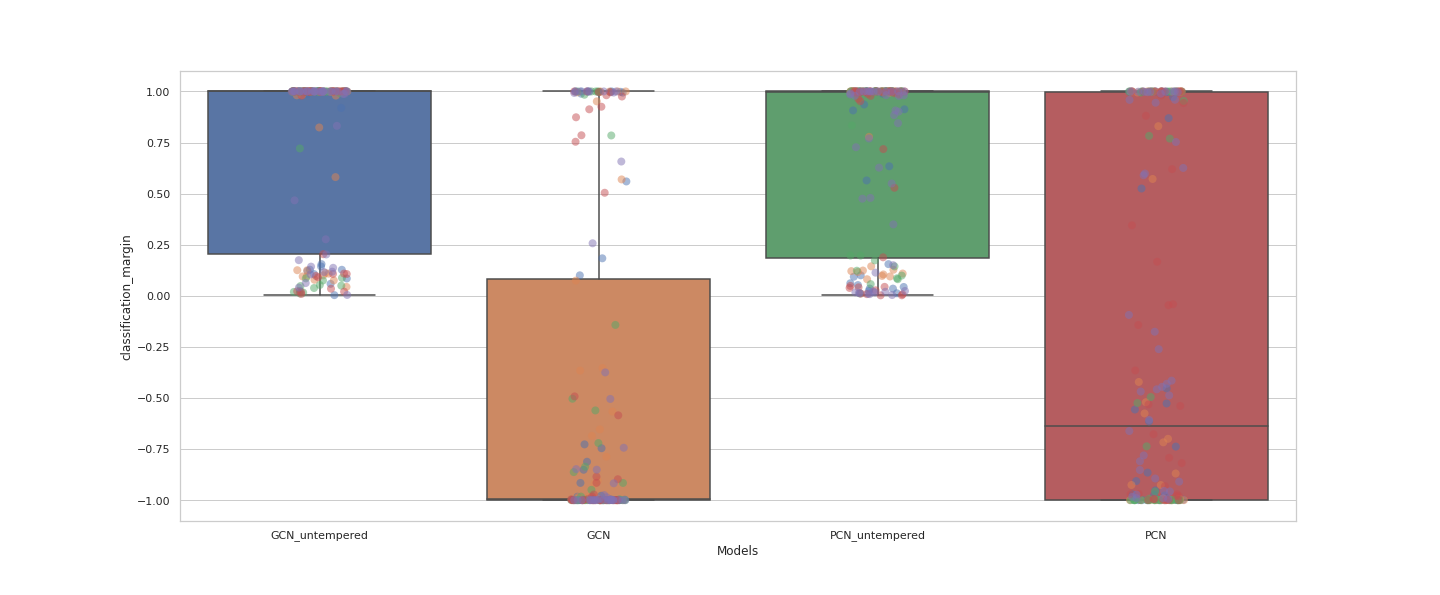}
    }
    \caption{Classification margin diagram of targeted attack on features when the perturbation rate is~50.}
    \label{fig:only_f_attack50}
\end{figure}

\begin{figure}[H]
    \centering
    \makebox[\textwidth]{
    \includegraphics[width=\textwidth]{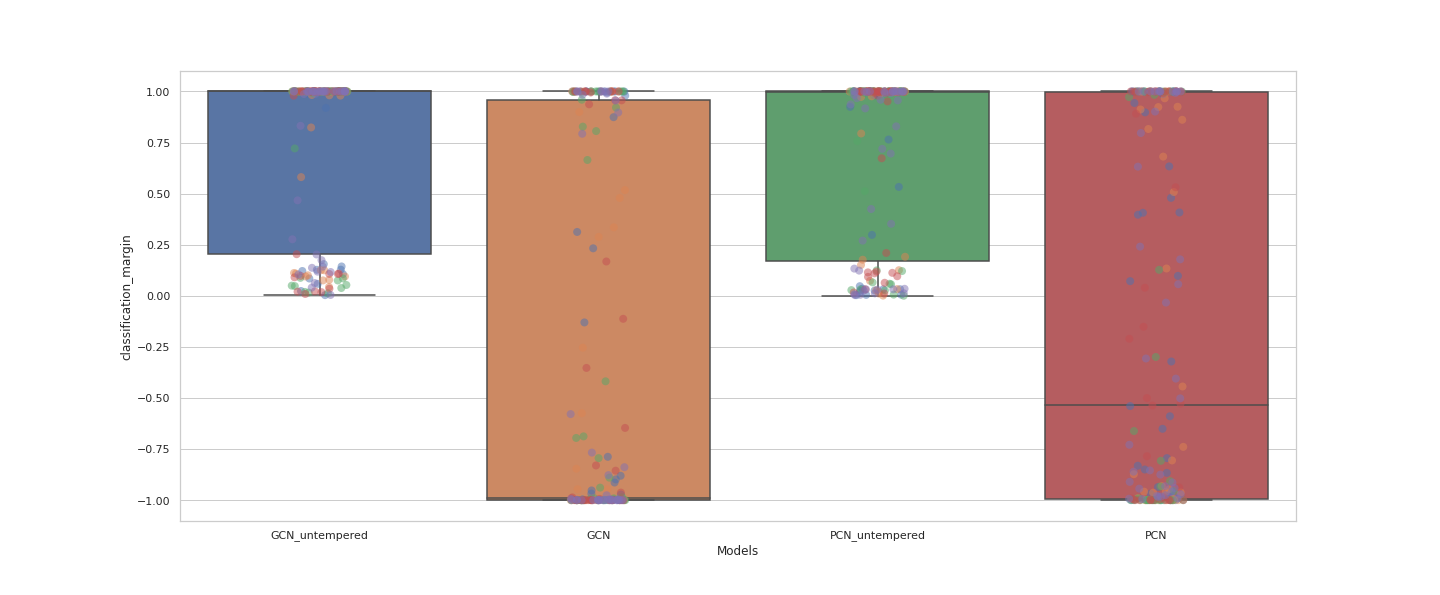}}
    \caption{Classification margin diagram of targeted attack on structure when the perturbation rate is~2.}
    \label{fig:only_f_attack2}
\end{figure}

\begin{figure}[H]
    \centering
    \makebox[\textwidth]{
    \includegraphics[width=\textwidth]{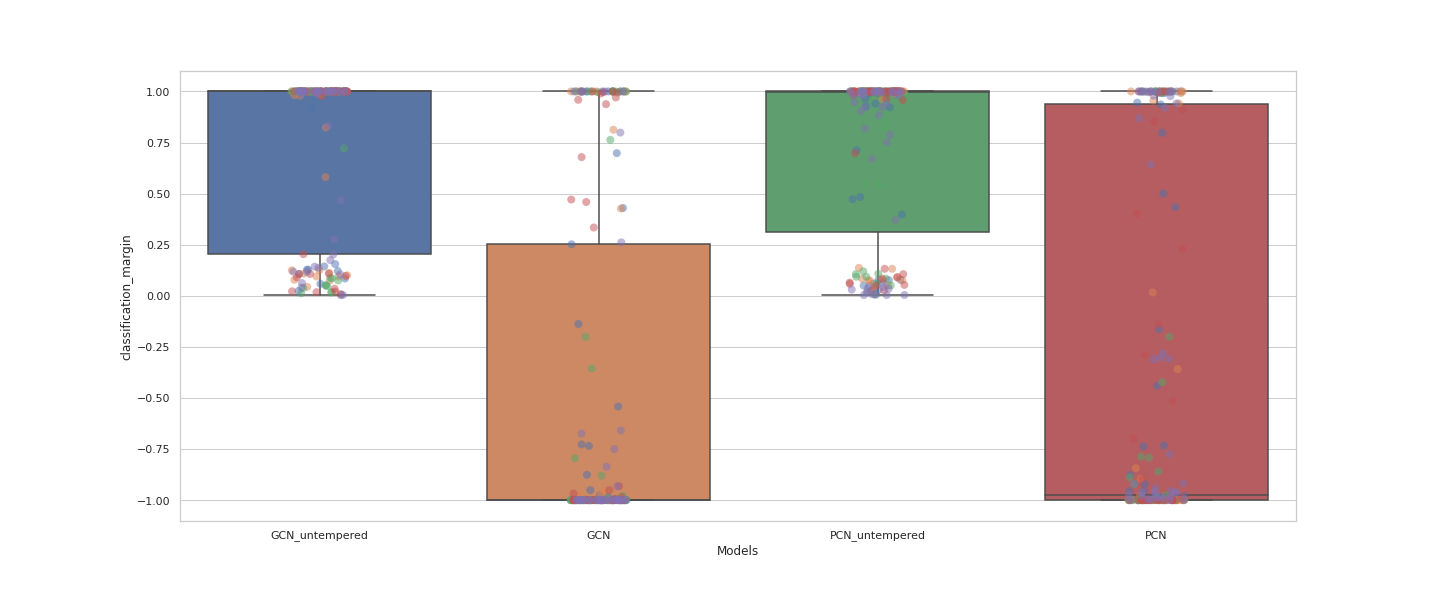}}
    \caption{Classification margin diagram of targeted attack on structure when the perturbation rate is~3.}
    \label{fig:only_f_attack3}
\end{figure}

\textbf{(3) Structure attacks}: GPCNs also consistently outperform GCNs under structure-only attacks on 1, 2, 5, and 10 perturbations, as it can be observed in Figs.~\ref{fig:only_f_attack1}, \ref{fig:only_f_attack2}, \ref{fig:only_f_attack3}, \ref{fig:only_f_attack5}, and \ref{fig:only_f_attack10}.

\begin{figure}[H]
    \centering
    \makebox[\textwidth]{
    \includegraphics[width=\textwidth]{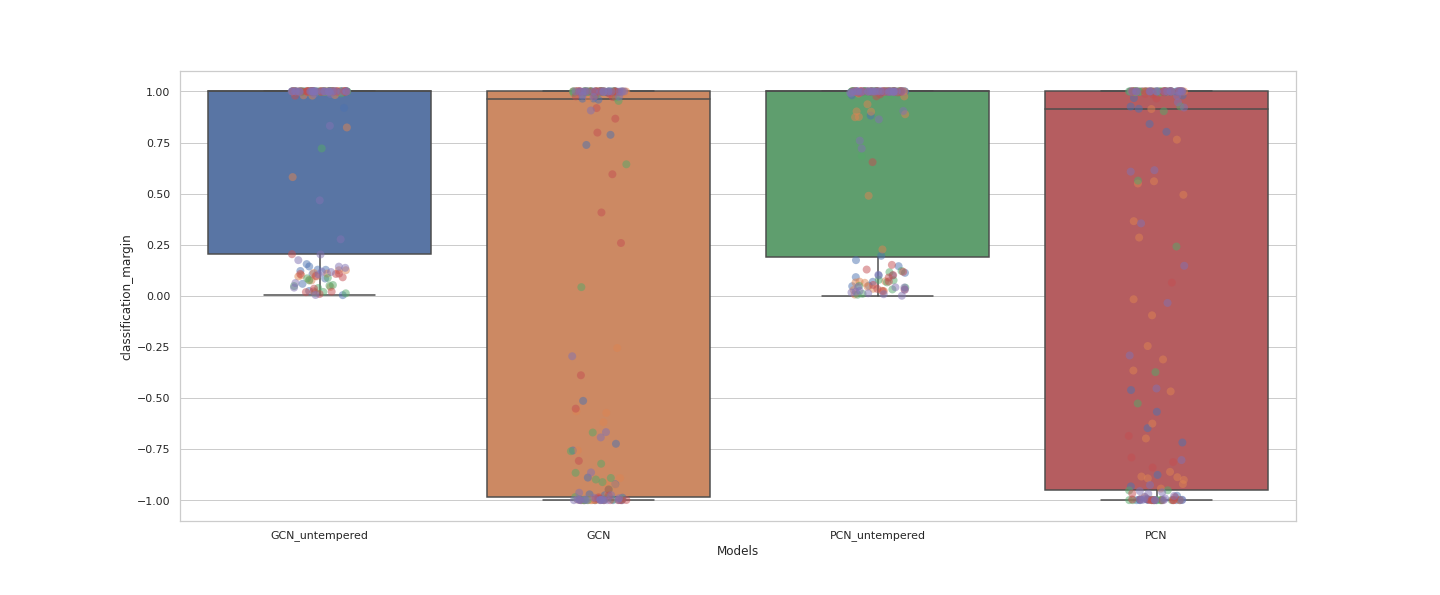}}
    \caption{Classification margin diagram of targeted attack on structure when the perturbation rate is~1.}
    \label{fig:only_s_attack1}
\end{figure}

\begin{figure}[H]
    \centering
    \makebox[\textwidth]{
    \includegraphics[width=\textwidth]{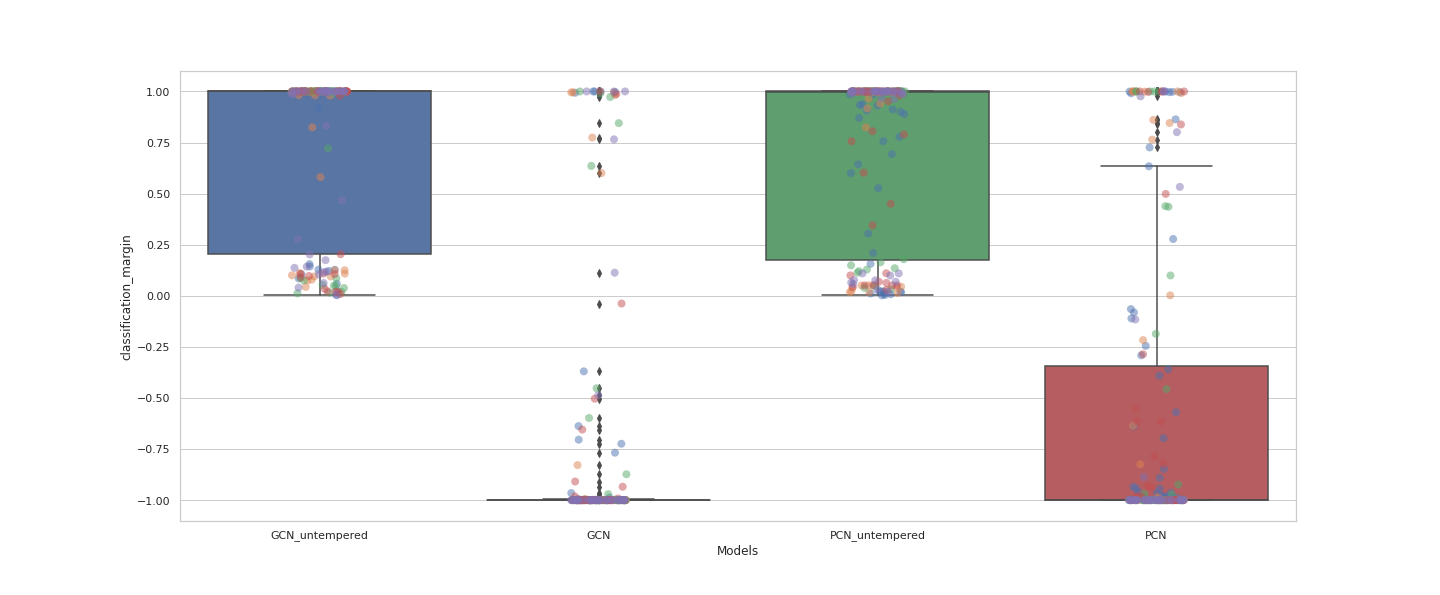}}
    \caption{Classification margin diagram of targeted attack on structure when the perturbation rate is~5.}
    \label{fig:only_s_attack5}
\end{figure}

\begin{figure}[H]
    \centering
    \makebox[\textwidth]{
    \includegraphics[width=\textwidth]{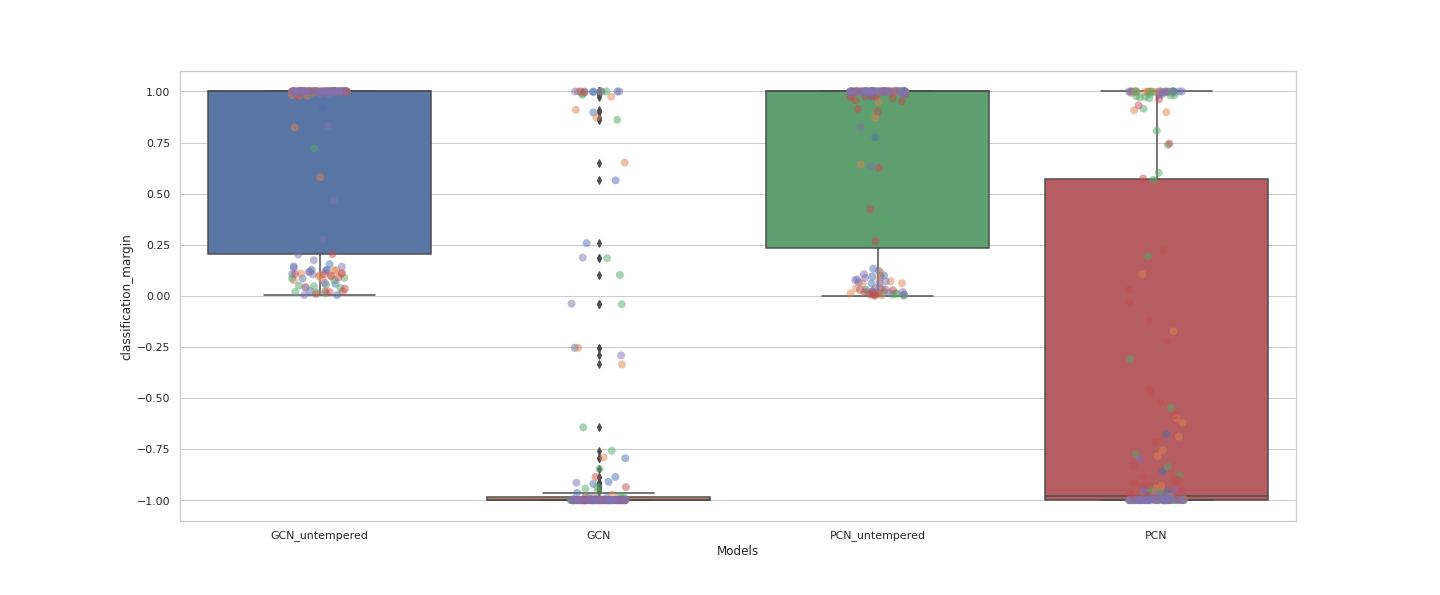}}
    \caption{Classification margin diagram of targeted attack on structure when the perturbation rate is~10.}
    \label{fig:only_s_attack10}
\end{figure}

\textbf{(4) Indirect attacks}:
For indirect attacks, we choose 5 influencing/neighboring nodes to attack for each victim node. We observe a similar trend that was found in the Nettack paper \citep{zugner2018adversarial}. We found that  indirect attacks do not affect GNNs as much as other attacks, as it can be witnessed from the box plots below. However, we also found that GPCNs, especially with smaller inference steps,  consistently outperform all models, but all GPCN models are strictly better than GCNs under all perturbations.

\begin{figure}[H]
    \centering
    \makebox[\textwidth]{
    \includegraphics[width=\textwidth]{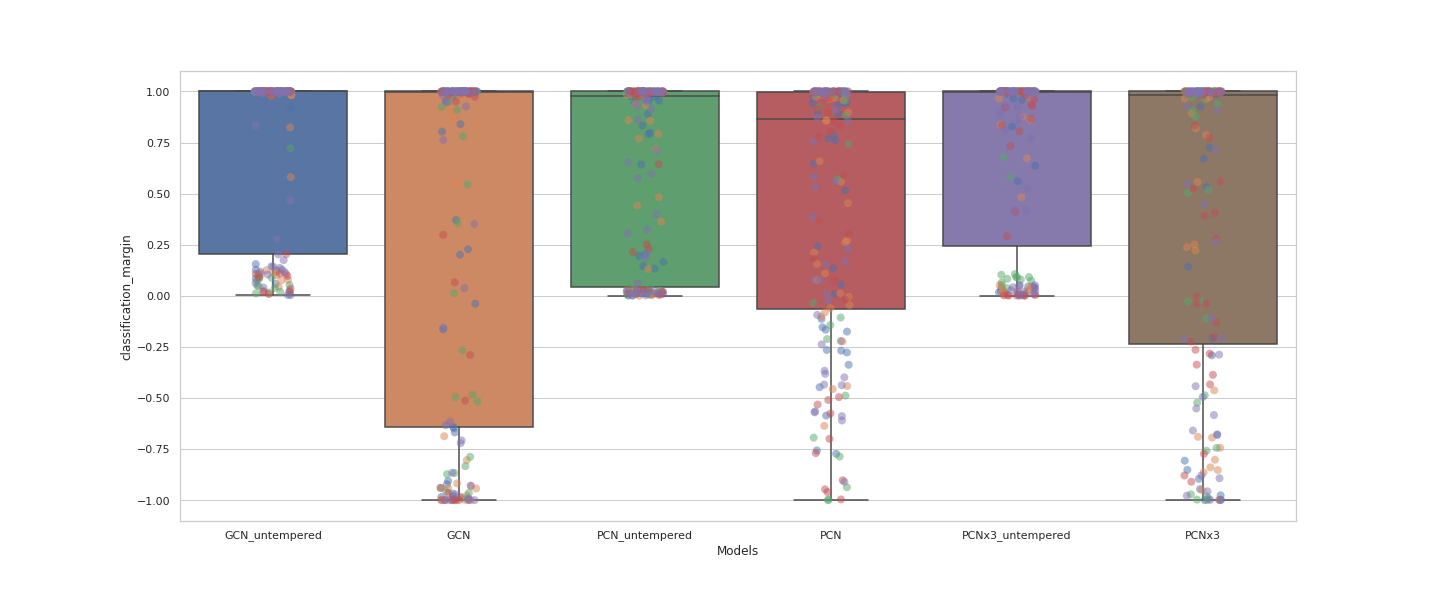}
    }
    \caption{Classification margin diagram on influence attack with perturbation equal to the degree of target node and 5 influencing neighboring nodes.}
    \label{fig:influence_attach_attack1}
\end{figure}

\begin{figure}[H]
    \centering
    \makebox[\textwidth]{
    \includegraphics[width=\textwidth]{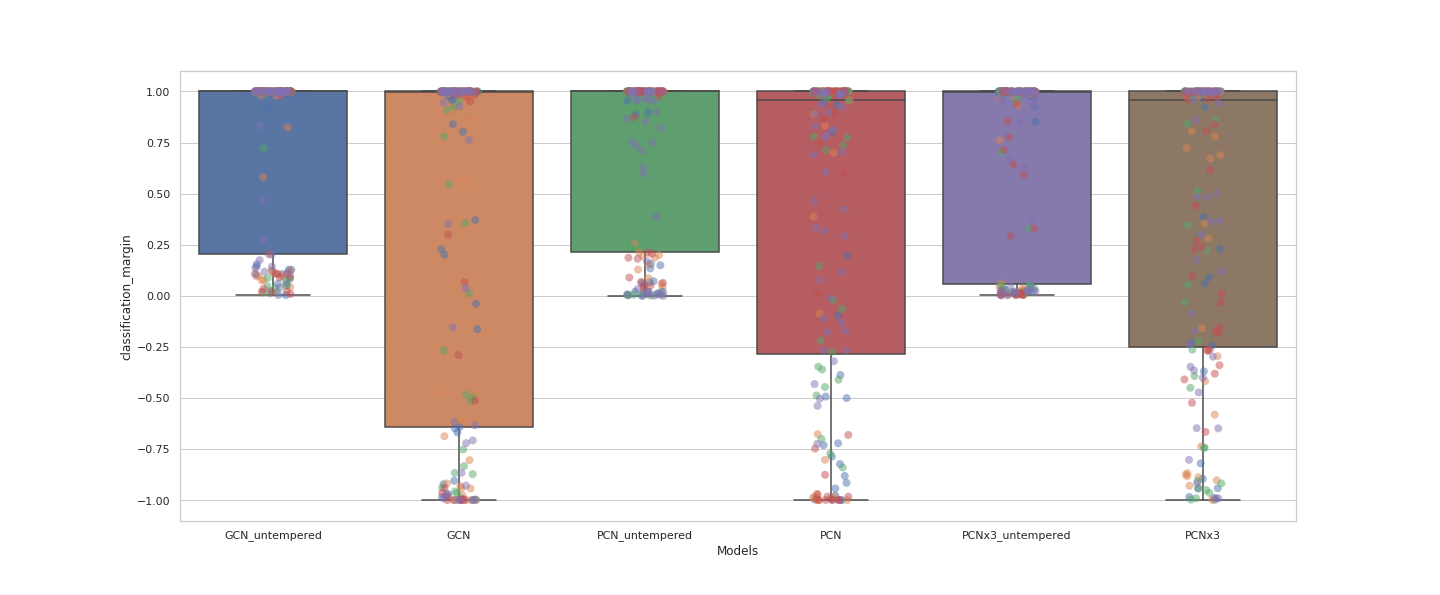}
    }
    \caption{Classification margin diagram on influence attack with perturbation equal to the degree of target node and 5 influencing neighboring nodes.}
    \label{fig:influence_attach_attack2}
\end{figure}

\begin{figure}[H]
    \centering
    \makebox[\textwidth]{
    \includegraphics[width=\textwidth]{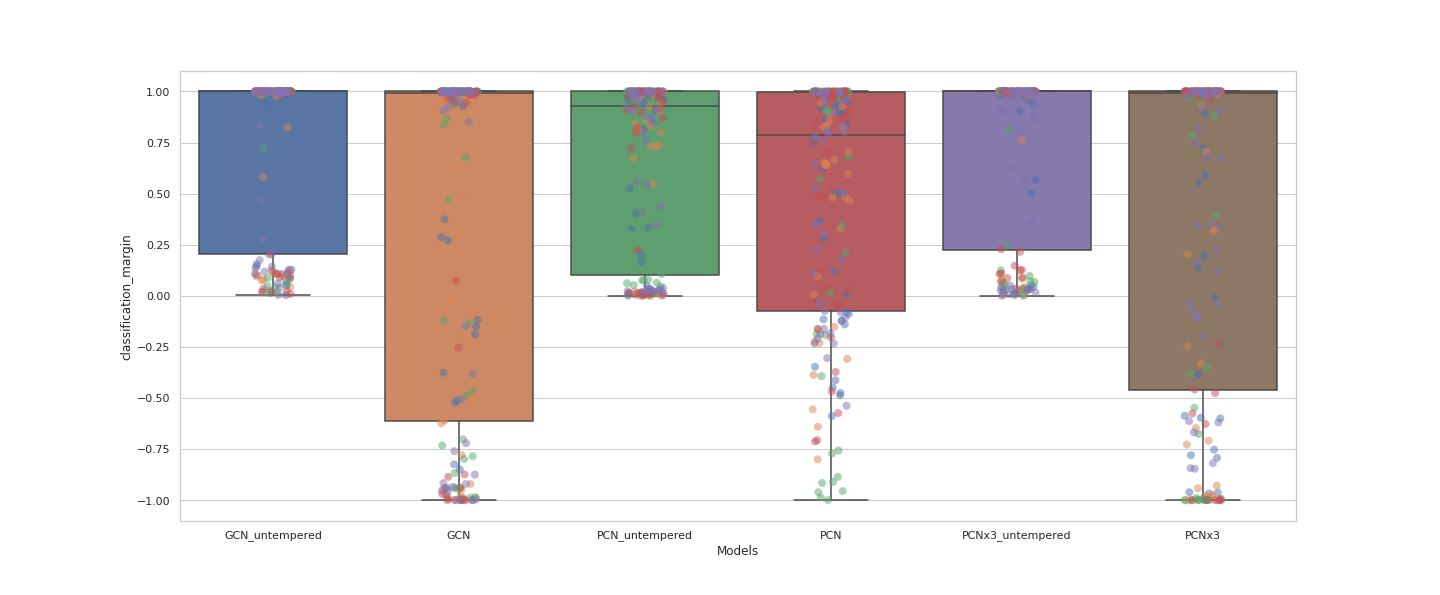}
    }
    \caption{Classification margin diagram on influence attack with perturbation equal to 1 and 5 influencing neighboring nodes.}
    \label{fig:influence_attach_attack3}
\end{figure}

Note that for Figs.~\ref{fig:influence_attach_attack1}, \ref{fig:influence_attach_attack2}, and \ref{fig:influence_attach_attack3}, on the x-axis, BP indicates the GCN model, PC denotes the GPCN model trained using  12 inference steps, and  PCx3 indicate the GPCN model trained using 36 inference steps, hence achieving a lower training energy. The suffix ``untempered'' indicates the performance of the model on a clean graph. To interpret the plots, a more robust model is one that retains  higher classification margins after the attacks.

\begin{figure}[H]
    \centering
    \makebox[\textwidth]{
    \includegraphics[width=\textwidth]{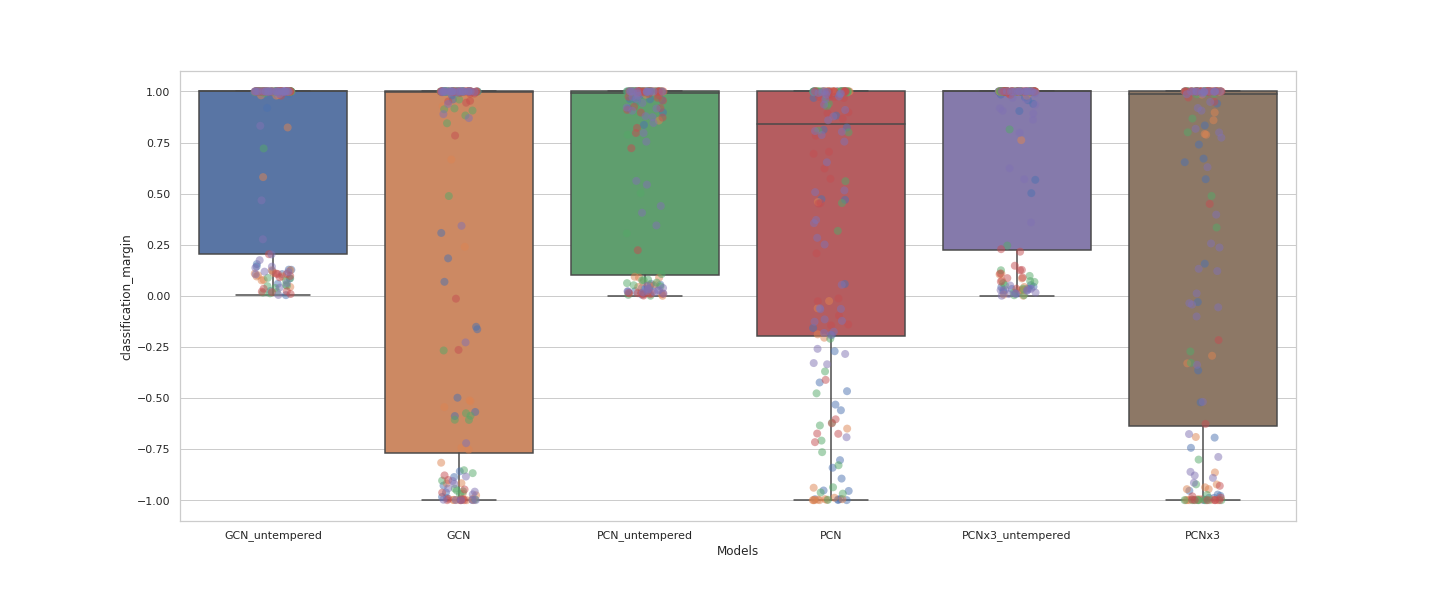}
    }
    \caption{Classification margin diagram on on influence attack with perturbation 10 and 5 influencing neighboring nodes.}
    \label{fig:influence_attach_attack4}
\end{figure}

\section{More Attacks}
\subsection{Global Poisoning  Attacks}
Following the same experimental setup as in Section~\ref{sec:mettack} for the global poisoning attack, we perform global random attacks \citep{jin2020graph},  which randomly insert fake edges into a graph, thus it can be viewed as adding a random noise to a clean graph. We evaluate our GPCN-GCN and GPCN-GAT against their similar architectural counterparts. Note that the results on the GAT model were taken from \citep{jin2020graph}, which had both batchnorm and dropout, unlike our GPCN-GAT. Table~\ref{table:random_poisoning} shows the results under different ratios of random noise from $0\%$ to $100\%$ with a step size of $20\%$. Each experiment is repeated five times under different random seeds, and we found that our GPCN-GCN and GPCN-GAT strictly outperform their counterparts, with GPCN outperforming all models on CORA and PubMed.

\textbf{Random Poisoning Attack}:
\ref{table:random_poisoning}

\begin{table}[H]
    \caption{Classification performance of the models under global poisoning random attack on structure. }
  \label{table:random_poisoning}
    
    \begin{adjustbox} {max width=\textwidth}

\begin{tabular}{ cc|ccccc }
\multicolumn{7}{c}{\textbf{\textit{Poisoning Attack with Random global attack on graph structure }}} \\
\hline
\textbf{Dataset} & \textbf{Ptb Rate (\%)} &  \textbf{GCN} & \textbf{GPCN-GCN} & \textbf{GPCN-GAT} & \textbf{GAT~\citep{jin2020graph}} & \textbf{RGCN~\citep{jin2020graph}} \\
\hline
     & $0$ & $82.87\pm0.75$   & ${83.17\pm0.78}$ &  $\mathbf{83.54\pm0.37}$ & $\mathbf{83.97\pm0.65}$ & $83.21\pm0.75$ \\
     & $20$ & $79.42\pm1.21$   & $\mathbf{81.06\pm0.58}$ & ${78.17\pm1.93}$ & ${77\pm0.5}$ & $80.12\pm0.24$ \\
CORA & $40$ & $76.46\pm1.36$   & ${78.8\pm0.48}$ & ${76.26\pm1.22}$ & ${75.4\pm0.57}$ & $\mathbf{79.05\pm0.55}$ \\
     & $60$ & $74.38\pm0.69$   & $\mathbf{77.56\pm0.65}$ & ${74.11\pm1.85}$ & ${71.5\pm1.59}$ & $77.01\pm0.98$ \\
     & $80$ & $72.33\pm1.16$   & $\mathbf{76.12\pm0.68}$ & ${72.44\pm0.91}$ & ${67.5\pm0.42}$ & $75.84\pm1.18$ \\
     & $100$ & $70.66\pm1.31$   & $\mathbf{74.34\pm1.14}$ & ${70.44\pm0.46}$ & ${65.7\pm0.12}$ & $72.10\pm0.23$ \\
\hline
         & $0$ & $72.43\pm0.05$   & ${72.64 \pm0.51}$ & $\mathbf{73.58\pm0.13}$ & ${73.26\pm0.83}$ & $70.97\pm0.37$\\
         & $20$ & $69.71\pm0.84$   & ${70.16\pm0.57}$ & $\mathbf{70.36\pm0.7}$  & ${69.2\pm0.59}$ & $67.50\pm0.71$\\
         & $40$ & $68.22\pm0.11$   & $68.28\pm1.29$ & $\mathbf{68.4\pm0.78}$ & ${65.4\pm0.54}$ & $63.28\pm0.62$\\
CiteSeer & $60$ & $66.55\pm0.59$  & $66.28\pm0.72$ & $\mathbf{66.72\pm0.97}$ & ${62.5\pm1.17}$ & $61.15\pm0.29$\\
         & $80$ &$65.26\pm0.96$   & $\mathbf{65.35\pm0.11}$ & ${64.34\pm0.58}$ & ${60.9\pm0.27}$ & $58.12\pm0.64$\\
         & $100$ & $63.76\pm1.12$   & $\mathbf{64.45\pm0.58}$ & ${63.88\pm0.42}$ & ${57.61\pm0.98}$ & $57.73\pm0.82$\\
\hline
       & $0$ & $85.37\pm0.06$   & $85.3\pm0.3$ & ${83.56\pm0.22}$ & ${83.73\pm0.40}$ & $\mathbf{86.09\pm1.61}$ \\
       & $20$ & $83.38\pm0.21$   & $\mathbf{83.05\pm0.13}$ & ${79.53\pm0.65}$ & ${79.1\pm0.25}$ & $\mathbf{83.82\pm0.98}$ \\
PubMed & $40$ & $81.84\pm0.22$  & $\mathbf{81.92\pm0.18}$ & ${77.4\pm0.58}$ & ${76.7\pm0.34}$ & $81.89\pm0.42$ \\
       & $60$ & $80.9\pm0.36$ & $\mathbf{80.79\pm0.46}$ & ${76.35\pm0.74}$ & ${74.8\pm0.59}$ & $\mathbf{80.95\pm1.22}$ \\
       & $80$ &$80.05\pm0.21$   & $\mathbf{80.09\pm0.18}$ & ${74.91\pm0.86}$ & ${73.3\pm0.74}$ & $79.36\pm0.71$ \\
       & $100$ & $79.18\pm0.27$   & $\mathbf{79.43\pm0.44}$ & ${73.52\pm0.97}$ & ${71.8\pm0.64}$ & $79.01\pm0.52$ \\

\end{tabular}

 \end{adjustbox}
 
\end{table}

\subsection{Evasion}

\textbf{Fast Gradient Sign Method (FGSM/FGA)}:

First, following the experimental setup in Section~\ref{sec:evasion}, we assess the robustness against structural evasion attacks, known as Fast Gradient Sign Method (FGSM/FGA) \citep{chen2020graph}. We randomly select $1000$ victim nodes from both the validation and the test set. As in previous works \citep{zugner2019adversarial, jin2020graph}, the perturbations budget ranges from 1 to 5, and each victim node is attacked separately. The results are shown in Table~\ref{table:fga}, where GPCNs strictly perform better than GCNs on both CORA and CiteSeer. Due to computation limitations, we test on PubMed and only using the initial Adam learning rate of 0.01, as it was done in multiple similar bodies of previous work \citep{zugner2019adversarial, jin2020graph}.

\begin{table}[H]
   
    \centering
 \caption{Classification performance of the models under evasion attack with FGA targeted attack on structure. }
  \label{table:fga}\begin{tabular}{cc | c c c }
\multicolumn{4}{c}{\textit{Evasion Attack with FGA targeted attack  on graph structure }} \\
\hline
\textbf{Dataset} & \textbf{$N^o$ of Perturbation}  & \textbf{GCN} & \textbf{GPCN-GCN} \\
\hline
     & $1$ & $73.0\pm5.14$ & $\mathbf{78.32\pm 5.73}$  \\
     & $2$ & $55.46\pm5.68$ & $\mathbf{57.44\pm6.00}$ \\
CORA & $3$ & $42.42\pm5.68$ & $42.32\pm4.97$ \\
     & $4$ & $34.16\pm4.54$ & $\mathbf{34.36\pm5.43}$ \\
     & $5$ & $26.92\pm4.19$ & $\mathbf{27.28\pm6.39}$ \\
\hline
     & $1$ & $71.56\pm6.32$ & $\mathbf{73.28\pm6.51}$  \\
     & $2$ & $52.36\pm5.49$ & $\mathbf{55.7\pm7.10}$ \\
CiteSeer & $3$ & $40.78\pm4.65$ & $\mathbf{44.1\pm7.72}$ \\
     & $4$ & $32.8\pm4.33$ & $\mathbf{37.82\pm9.29}$ \\
     & $5$ & $27.12\pm3.33$ & $\mathbf{31.52\pm5.17}$ \\
\hline

\end{tabular}

\end{table}

\end{document}